\documentclass[journal]{IEEEtran}

\ifCLASSINFOpdf
\usepackage[pdftex]{graphicx}
\else
\fi
\usepackage{amsmath}
\usepackage{algorithm}
\usepackage{algorithmic}
\usepackage[caption=false,font=footnotesize]{subfig}
\usepackage{url}

\begin{document}
%
\title{Fuzzy Semantic Segmentation of Breast Ultrasound Image with Breast Anatomy Constraints}
%
%
%

\author{Kuan~Huang,
        Yingtao~Zhang,
        H. D.~Cheng$^\dagger$,
        Ping~Xing,
        and~Boyu~Zhang
\thanks{K. Huang, H. D. Cheng, and B. Zhang are with the Department of Computer Science, Utah State University, Logan, UT 84341 USA, (e-mail: kuan.huang@aggiemail.usu.edu; hengda.cheng@usu.edu; boyu.zhang@aggiemail.usu.edu).}
\thanks{Y. Zhang is with the School of Computer Science and Technology, Harbin Institute of Technology, Harbin, China, (e-mail: yingtao@hit.edu.cn).}
\thanks{P. Xing is with the Ultrasound Department, the First Affiliated Hospital of Harbin Medical University, Harbin, China, (e-mail: shuijingping126@126.com).}
\thanks{${^\dagger}$Corresponding author: H. D. Cheng}}
\maketitle

\begin{abstract}
Breast cancer is one of the most serious disease affecting women's health. Due to low cost, portable, no radiation, and high efficiency, breast ultrasound (BUS) imaging is the most popular approach for diagnosing early breast cancer. However, ultrasound images are low resolution and poor quality. Thus, developing accurate detection system is a challenging task. In this paper, we propose a fully automatic segmentation algorithm consisting of two parts: fuzzy fully convolutional network and accurately fine-tuning post-processing based on breast anatomy constraints. In the first part, the image is preprocessed by contrast enhancement, and wavelet features are employed for image augmentation. A fuzzy membership function transforms the augmented BUS images into fuzzy domain. The features from convolutional layers are processed using fuzzy logic as well. The conditional random fields (CRFs) post-process the segmentation result. The location relation among the breast anatomy layers is utilized to improve the performance. The proposed method is applied to the dataset with 325 BUS images, and achieves state-of-art performance compared with that of existing methods with true positive rate 90.33\%, false positive rate 9.00\%, and intersection over union (IoU) 81.29\% on tumor category, and overall intersection over union (mIoU) 80.47\% over five categories: fat layer, mammary layer, muscle layer, background, and tumor.
\end{abstract}


%
\IEEEpeerreviewmaketitle

\section{Introduction}
%
%
%
%
\IEEEPARstart{B}{reast} cancer occurs frequently in women over the world. It is the most serious disease and the second common cancer after skin cancer among women in the United States until 2019 \cite{ref1,ref2}. The most urgent is to diagnose breast cancer in early stage. It lacks obvious symptom in early stage of breast cancer; therefore, many patients miss the best chance to cure it. Investigated by some organizations, the survival rate of stages 0 and 1 of breast cancer during 2007 and 2013 is close to 100\% \cite{ref3}. Breast ultrasound (BUS) imaging is harmless, low cost, portable and effective; therefore, it becomes the most important approach for breast cancer early detection. However, breast ultrasound (BUS) images usually have poor quality, low contrast and large uncertainty. The main causes of the uncertainty are: 1) the BUS images are acquired from different machines during different periods with different settings; 2) the characteristics of breasts of different people might be various; and 3) the resolution of BUS image is low, and it has inherent noise, speckles. In addition, the quality of the BUS images depends on the skills and the ways of the radiologists to acquire the images. 

In order to prevent misdiagnosis, computer-aided diagnosis (CAD) is studied extensively. Image segmentation and classification are two most important components in CAD systems \cite{ref4}. BUS image segmentation consists of semi-automatic and fully-automatic methods \cite{ref30,ref36}; and it can be thresholding methods \cite{ref5,ref6}, clustering-based methods \cite{ref7,ref8}, watershed-based methods, graph-based methods \cite{ref32}, active-contour model \cite{ref9}, Markov model \cite{ref42}, neural network \cite{ref43}, deep learning \cite{ref13,ref16} etc.

Gray level thresholding method and area growing lesion contour detecting method were studied in \cite{ref5,ref6}. The region of interest (ROI) was determined by thresholding, and then a maximization utility function was applied to ROI for obtaining the lesion contour. The utility function was the average radial derivative (ARD), which calculated the radial direction derivatives from the seed point to the boundary. The seed point of each image was chosen as the center of ROI. The dataset contained 400 cases (757 images). The area under receiver operating characteristic (ROC) curve ($A_z$) was utilized to evaluate the performance which was 0.91.

Moon et al. proposed a clustering-based breast cancer segmentation method \cite{ref7}. The method consisted of three parts. The first part was quantitative tissue clustering. The tissue within tumor is different from other tissues. A 3-D mean shift clustering was used for selecting tumor tissues according to the echogenicities. The fuzzy c-means clustering method divided the segmented regions into four clusters. The morphology and echogenicity features were extracted, and logistic regression was used to classify the benign and malignant tumors.

Shan et al. proposed a fully automatic breast cancer segmentation method based on neutosophic I-means clustering \cite{ref8}. It used an automatic seed point selection algorithm to generate the ROI and proposed a novel contrast enhancement method based on frequency and spatial domain. A clustering method combined with neutrosophic logic, the neutosophic I-means clustering, was developed to segment BUS image. The method achieved true positive rate (TPR) of 92.4\%, false positive rate (FPR) of 7.2\%, and similarity rate of 86.3\%; mean Hausdorff error (AHE) of 22.5 pixels and mean absolute error (AME) of 4.8 pixels, respectively.

Xian et al. \cite{ref17} developed a fully automatic segmentation method based on the characteristics of BUS image in frequency and spatial domain. The method had two parts: fully automatic breast tumor ROI generation, and a robust tumor segmentation based on ROI. In ROI generation step, a fully automatic reference point selection method was designed using breast anatomy: the location of the tumor was in the middle of the pre-mammary layer and retro-mammary layer. The mean shift algorithm was utilized to extend the reference points as the seed points. Finally, the ROI of tumor was calculated by using the seed points. In tumor segmentation step, a minimization cost function was used, and the frequency and spatial boundary information was utilized as the constraint of the cost function. The approach achieved 91.23\% TPR, 9.97\% FPR, and similarity rate of 83.73\% using a dataset of 184 images.

The deep neural networks have been utilized for image segmentation and classification. In \cite{ref10,ref11,ref12}, and deep networks are proposed for breast histology image and mammographic mass segmentation. A deep learning approach was applied for breast cancer detection \cite{ref13}. Three network structures were used: patch-based LeNet \cite{ref41}, U-Net \cite{ref14}, and transfer learning with a pretrained fully convolutional network (FCN) with AlexNet \cite{ref15}. The network structures were applied to two datasets with 469 images. The experiments were conducted to compare three network structures: 1) trained on a dataset and tested on another; 2) trained and tested on single dataset; 3) trained on the combined dataset and tested on individual one. In the first experiment, the result on dataset A using U-Net was TPR 0.83, FPs/Image 0.08, and F-measure 0.87; the result on dataset B was TPR 0.70, FPs/Image 0.66, and F-measure 0.59. The result of the second experiment was TPR 0.98, FPs/Image 0.16, and F-measure 0.91 on dataset A using FCN-AlexNet; and TPR 0.92, FPs/Image 0.17, and F-measure 0.89 on dataset B. The result of the third experiment was TPR 0.99, FPs/Image 0.16, and F-measure 0.92 on dataset A using FCN-AlexNet; and TPR 0.93, FPs/Image 0.18, and F-measure 0.88 on dataset B. It concluded that the performance depended on dataset.

BUS image semantic segmentation was studied in \cite{ref16}. An extracted feature of 1-level wavelet transform was added to the input of the neural network. The prior knowledge that tumor must be inside mammary layer was utilized to constrain the conditional random fields (CRFs) energy function. This method could detect and segment the mammary layer and tumor; while other layers were treated as background. The method achieved TPR 92.80\%, FPR 9.00\%, and Intersection over Union (IoU) 82.11\% in tumor segmentation.

According to above discussion, we can summarize that although all existing methods claimed to achieve the best results on their own datasets, they still have shortcomings: 1) traditional segmentation methods were based on the assumption that there was one and only one tumor in each image. Hence, the methods could not handle the situation: there is no tumor or more than one tumor. 2) most of the segmentation methods did not employ breast anatomy knowledge. 3) the uncertainty in BUS images, especially around the boundaries between breast anatomy layers, is not handled. Due to the low resolution and poor quality of the BUS image, the boundary areas were very vaguer and fuzzy, even difficult for doctors to classify.

To overcome the shortcomings, a novel, robust, fuzzy logic guided BUS image semantic segmentation method with breast anatomy constrained post-processing method is proposed. Based on \cite{ref18}, an adaptive membership function is designed. The input image is transformed to fuzzy domain using the membership function. The fuzzy features are used to represent the uncertainty in the image. Fuzziness of the image is reduced by multiplying uncertainty maps and input image. The feature maps of the first convolutional layer are transformed in fuzzy domain and uncertainty in the convolutional feature maps is reduced by multiplying uncertainty maps and the corresponding convolutional feature maps. Then, the context information (breast anatomy) and the layer structure of human breast \cite{ref20} are applied to the conditional random fields (CRFs) to improve the final segmentation result. The dataset contains 325 BUS images labeled by the doctors of cooperative hospitals; 229 images contain tumors, and the rest ones do not contain tumors. Previously existing datasets just had labels for tumors, while our dataset has the information of 5 categories: fat layer, mammary layer, muscle layer, background, and tumor in every image (as shown in Fig. 1).

The paper is organized as follows: in Section 2, the fuzzy information guided fully convolutional network is introduced; in Section 3, the breast anatomy constrained CRFs are presented; in Section 4, the experiments for the network structure and post processing are discussed, the performance is compared with that of previous methods, and the conclusions are given in Section 5.
\begin{figure}[tbp]
    \centering
	\subfloat[]{\includegraphics[width=4cm,height=3cm]{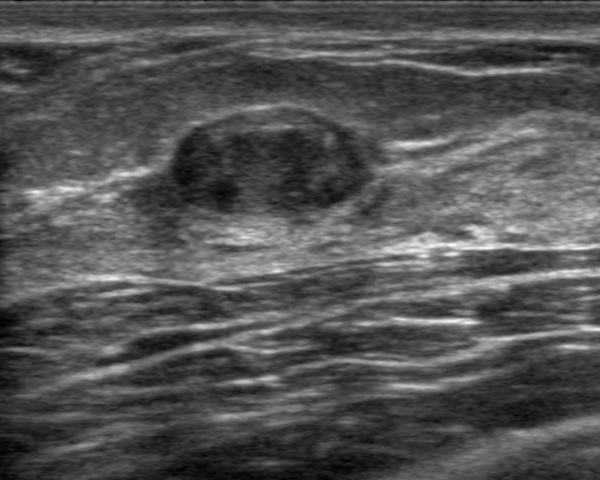}}\quad
	\subfloat[]{\includegraphics[width=4cm,height=3cm]{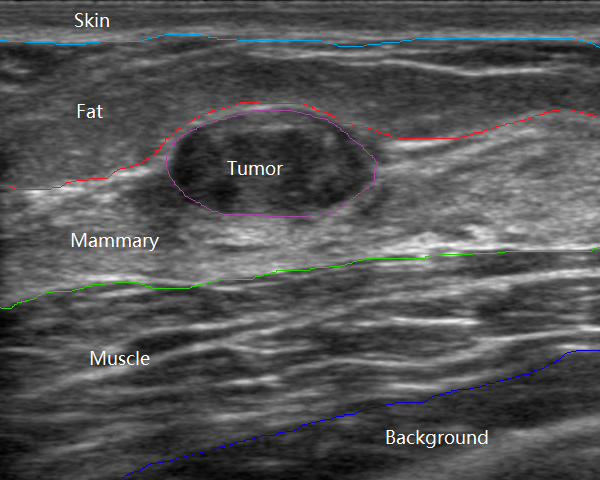}}\\
	\captionsetup{justification=centering}
	\caption{(a) Breast ultrasound image; (b) breast anatomy.}
\end{figure}

\section{Fuzzy information guided fully convolutional network}
Fully convolutional network (FCN) \cite{ref21} was widely used in semantic segmentation because FCN obtained better results than that of traditional methods. FCN allowed to use the whole images as inputs and provided a pixel-wise category prediction. Supervised learning methods using deep convolutional neural network for image analysis can be categorized as: 1) patch-based methods: input images were divided into small overlapping or non-overlapping patches which were classified. They represented the category of the center pixel or a chosen pixel on the boundary of a patch. A sliding window was used to scan all the pixels in the image. 2) FCN: FCN realized an end-to-end segmentation, where the whole images were utilized as the inputs without dividing into patches. FCN performed better than patch-based methods and was applied to nature image semantic segmentation, medical image processing, crack detection, and other tasks. In this research, a fuzzy FCN is proposed for BUS image segmentation.

Ideal fully automatic BUS image segmentation system should have the following characteristics: 1) it uses the entire image as the input, and the output is the final segmentation result without manual intervention; 2) it has high robustness and accuracy. In Fig. 2, (a) is the original image, and (b) is the label map created by experienced doctors: the black areas represent background, the green area represents fat layer, the yellow area represents mammary layer, the blue area represents muscle layer, and the red area represents tumor. In Fig. 2 (a) the regions marked by red and green rectangles are hard to determine even by experienced radiologists, since the boundaries are vaguer, fuzzy and with high uncertainty.

A membership function transforms BUS image to fuzzy domain. The uncertainty in BUS image can be handled well by using fuzzy logic, and better semantic segmentation result can be obtained.
\begin{figure}[tbp]
    \centering
	\subfloat[]{\includegraphics[width=4cm,height=3cm]{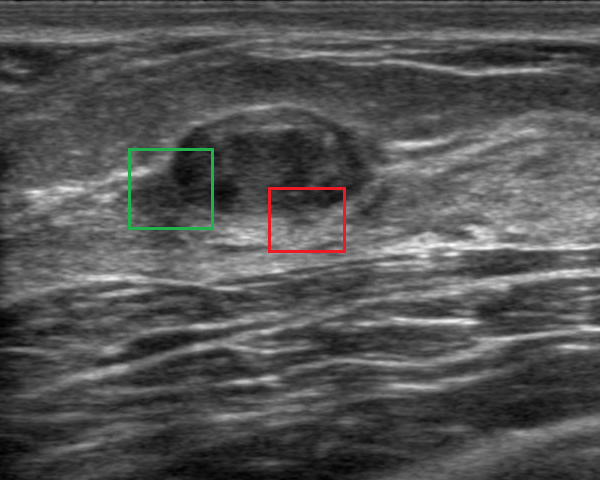}}\quad
	\subfloat[]{\includegraphics[width=4cm,height=3cm]{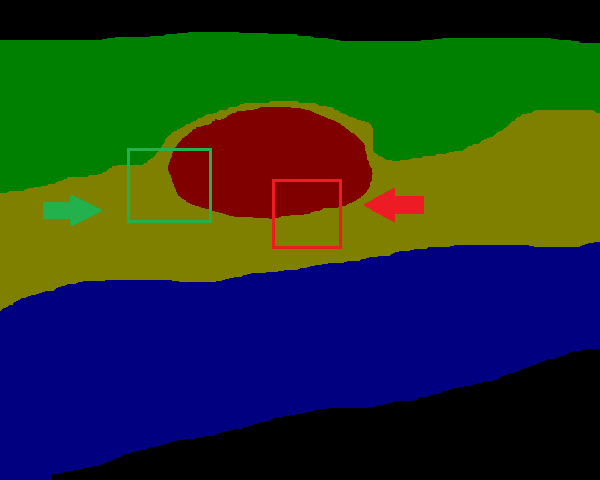}}\\
	\caption{Uncertainty in BUS image: (a) BUS image and uncertain areas; (b) the label map of the BUS image.}
\end{figure}
\subsection{Overview}
The flowcharts of the proposed approaches are shown in Fig. 3. In Fig. 3 (a), the input image is preprocessed by contrast enhancement. Then, wavelet transform is applied. The original image and wavelet information are transformed to fuzzy domain by membership functions to deal with the uncertainty. Results after reducing the uncertainty are input into the first convolutional layer. The obtained feature maps are transformed into fuzzy domain as well, and the uncertainty is reduced by multiplying the uncertainty maps and the corresponding feature maps. In Fig. 3 (b), wavelet transform (red part in Fig.3 (a)) is not utilized. After reducing uncertainty in gray-level intensity and the first convolutional layer, the network can achieve similar performance to that of Fig. 3 (a). Two approaches are evaluated by segmentation accuracies and compared with the original fully convolutional network.
\begin{figure*}[tbp]
    \centering
	\subfloat[]{\includegraphics[scale = 0.45]{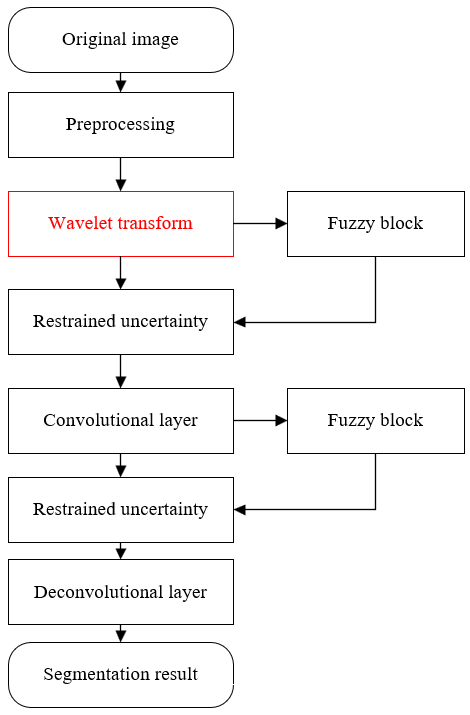}}\hspace{5em}
	\subfloat[]{\includegraphics[scale = 0.5]{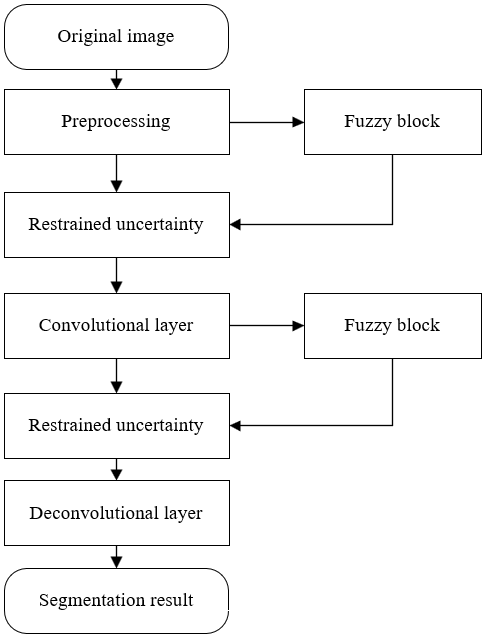}}\\
    \caption{Flowchart of the two strategies of the fuzzy FCNs: (a) using wavelet; (b) without using wavelet.}
\end{figure*}
\subsection{Preprocessing}
\textbf{Histogram equalization:} The original images were captured in different periods which might have different ranges of intensities. It will affect the segmentation results. The histogram equalization is modified to make the input image have the intensity range from 0 to 255, and to conduct contrast enhancement. Histogram equalization is performed on both training set and testing set. In histogram equalization, the probability of a pixel with intensity $\theta$, $p_z (\theta)$, is computed by Eq. (1) \cite{ref40}
\begin{equation}
p_z (\theta) = p(z=\theta)= \frac{n_\theta}{n}, \, 0 \leq \theta \leq L_z-1
\end{equation}
where $n_\theta$ represents the number of pixels with intensity $\theta$; $n$ represents the total number of pixels. $L_z$ is the largest intensity. The cumulative distribution function of $p_z (\theta)$ is defined as: \par
\begin{equation}
cdf_z (\theta) = \sum_{u=0}^\theta {p_z (\theta)}
\end{equation}

The new intensity $h(\theta)$ is computed by
\begin{equation}
h(\theta) = \left \lfloor \frac{cdf_z (\theta) - cdf_{zmin}}{1 - cdf_{zmin} } \times 255 \right \rfloor 
\end{equation}
where $\theta$ represents the original intensity, and $cdf_{zmin}$ is the minimum non-zero value in the cumulative distribution function.

The original images are shown in Fig. 4 (a). After histogram equalization, the contrast is enhanced, and the range of the intensities is normalized from 0 to 255 as shown in Fig. 4 (b).

\textbf{Wavelet transform:} to overcome small dataset size problem, high-pass filter \textit{H} and low-pass filter \textit{G} of wavelet transform are used to obtain the high frequency and low frequency information. In this research, one level Haar wavelet transformation is applied, and the input image becomes a 3-channel image. The first channel is the original image; the second channel contains the low frequency coefficients; and the third channel contains the high frequency coefficients. Fig. 5 shows the original images and augmented 3-channel images, respectively.
\begin{figure}[tbp]
    \centerline{\includegraphics[width=8cm,height=6cm]{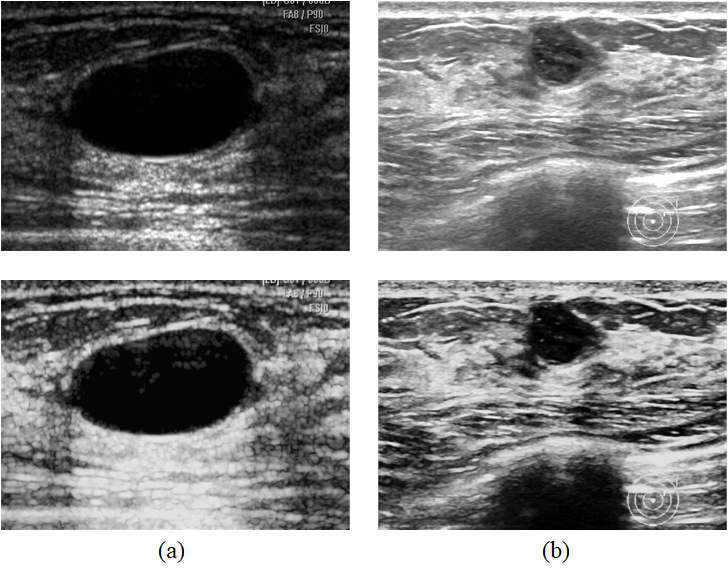}}
    \caption{(a) Original images; (b) images after histogram equalization.}
\end{figure}
\begin{figure}[tbp]
    \centerline{\includegraphics[width=8cm,height=6cm]{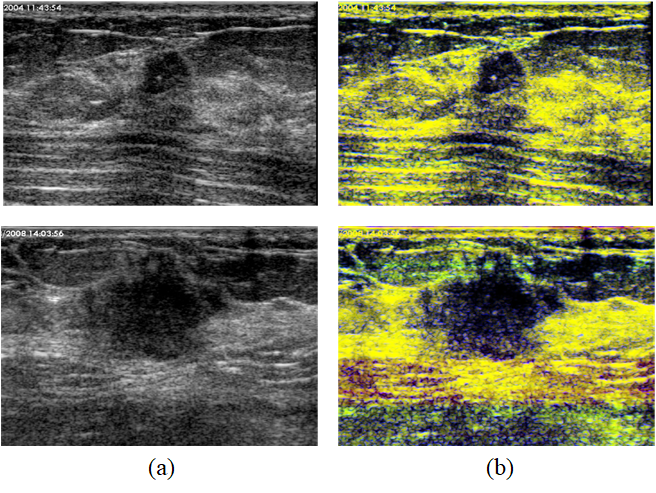}}
    \caption{(a) Original images; (b) augmented 3-channel images}
\end{figure}
\subsection{Fuzzy layer}
\begin{figure}[tbp]
    \centerline{\includegraphics[width=8cm,height=7cm]{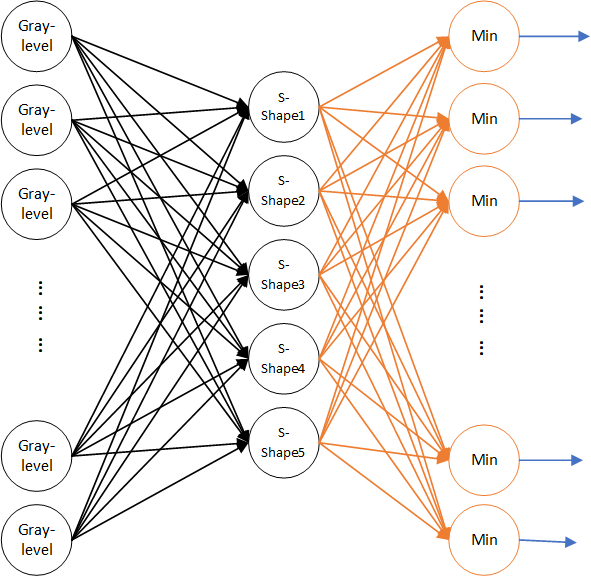}}
    \caption{Fuzzy layer structure.}
\end{figure}
The boundary between the layers is hard to determine due to the uncertainty, poor contrast, and inherent noise (speckles) in BUS image. Fuzzy logic has been applied to handle the uncertainty successfully. A fuzzy contrast enhancement method was developed \cite{ref23}. The membership function was a S-function with adaptive parameters calculated by using the local maxima of the histogram. A fuzzy clustering method was utilized for image segmentation \cite{ref24}. The fuzzy membership was initialized by k-means clustering. The segmentation cost function was based on the membership of each pixel and the Euclidean distances from the pixels to the cluster center. An edge-detection method based on generalized type-2 fuzzy logic was designed \cite{ref25}. The membership function was defined using Gaussian generalized type-2 membership functions. Fuzzy image processing methods can obtain robust results and handle uncertainty and noise well. In this research, we apply fuzzy logic to FCN to solve the uncertainty. The fuzzy fully convolutional network consists of three parts: 1) fuzzification layer; 2) uncertainty representation layer, and 3) fusion of convolutional network and fuzzy information. 

\textbf{Fuzzification layer:} Traditional fuzzification uses membership functions such as S-shape membership function, Gaussian membership function, Triangular membership function, etc. Parameters in the membership functions are decided manually or calculated using the information of the specific problems. In this paper, a trainable membership function is utilized to improve robustness and rationality.

In Fig. 6 (black part), the input images are transformed into fuzzy domain. Two membership functions are employed: trainable Sigmoid and Gaussian membership functions. Each input node (pixel) is transformed by the membership function. Let $x_i$ be the input node, \textit{i} represents the \textit{i}th pixel. All the input channels will conduct fuzzy transform. Here the gray-level channel is used as an example to show the membership and uncertainty intuitively. $o_i^r$ represents the output node and \textit{r} represents the category index. The trainable Sigmoid membership function for fuzzification layer is computed by Eq. (4)
\begin{equation}
o_i^r = \frac{1}{1+e^{a_i^r (x_i - b_i^r)}}, i=1,2,3,...,n;r=0,1,2,3,4    
\end{equation}
where $n$ represents the number of pixels in the image; \textit{r} has 5 values: 0 represents the background; 1 represents the tumor; 2 represents the fat layer; 3 represents the mammary layer; and 4 represents the muscle layer. $a_i^r$  and $b_i^r$ represent the parameters of the membership function for pixel \textit{i}. For every category, the pair of parameters are obtained during training, and the membership of the category is calculated using these parameters. In BUS images, tumor areas have low intensities in spatial domain, and other layers have higher intensities. By changing the parameters $a_i^r$  and $b_i^r$, trainable Sigmoid function can represent the membership of each category. The trainable Gaussian membership function is also used to compare with trainable Sigmoid membership function to demonstrate the usefulness of the fuzzy logic in handling uncertainty. The trainable Gaussian membership function is computed by Eq. (5)
\begin{equation}
    o_i^r = e^{- \frac{(x_i - \mu_i^r)^2} {2 {\sigma_i^r}^2}}, \, i=1,2,3,...,n;r=0,1,2,3,4
\end{equation}
where $\mu_i^r$ and ${\sigma_i^r}^2$ represent the mean and variance of category \textit{r}, which are utilized to obtain the memberships of different categories.

The fuzzy memberships are normalized by Eq. (6). It makes the summation of memberships in different categories of a pixel become one.\par
\begin{equation}
    o_i^r = \frac{x_i^r}{\sum_{r=0}^4 x_i^r}
\end{equation}

Heatmaps in Fig. 7 represent the membership values of the intensities; blue represents low membership value and red represents high membership value. In Fig. 7 (a)-(e), the memberships are computed by the trainable Gaussian membership function; and in (f)-(j), the memberships are computed by the trainable Sigmoid membership function.\par
\begin{figure*}[tbp]
    \centerline{\includegraphics[width=18cm,height=5cm]{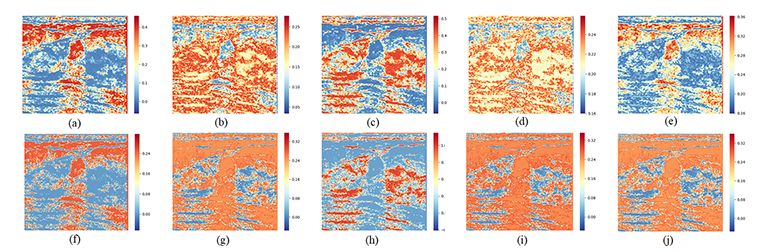}}
    \caption{The heatmaps of the membership maps: in (a)-(e) the memberships of tumor, fat layer, mammary layer, muscle layer, and background computed by the trainable Gaussian function; (f)-(j) the corresponding memberships computed by the trainable Sigmoid function. Blue represents low membership value and red represents high membership value}
\end{figure*}
The parameter $b_i^r$ in trainable Sigmoid membership function is initialized by the mean of the intensities of all training samples in category \textit{r}. The parameter $a_i^r$ is initialized by the uniform distribution. The parameter $\mu_i^r$ is initialized by the mean of the intensities of all training samples in category \textit{r}, and ${\sigma_i^r}^2$ is initialized by the variance of the samples in the same category. All the input channels (original gray-level intensity and wavelet coefficients) are transformed to fuzzy domain.

\textbf{Uncertainty representation layer (Orange area in Fig. 6):} If the membership of a pixel is close to 1 or 0, the uncertainty of the pixel is low. If the membership is around 0.5, the uncertainty is high, and it is hard to determine which category the pixel belongs to. The inputs of this layer are the fuzzy memberships, and the uncertainties in corresponding categories are computed using Eq. (7).
\begin{equation}
o_i^r = \left\{
        \begin{aligned}
        2 \times x_i^r & & if \, x_i^r < 0.5, \, r = 0,1,2,3,4  \\
        2 \times (1-x_i^r) & & if \, x_i^r > 0.5, \, r = 0,1,2,3,4
        \end{aligned}
        \right.
\end{equation}
where $x_i^r$ is the membership of pixel \textit{i} in the \textit{r}th category, which is the output of the fuzzification layer. The heatmaps of the uncertainty maps on gray-level intensities are generated as shown in Fig. 8.\par
\begin{figure*}[tbp]
    \centerline{\includegraphics[width=18cm,height=5cm]{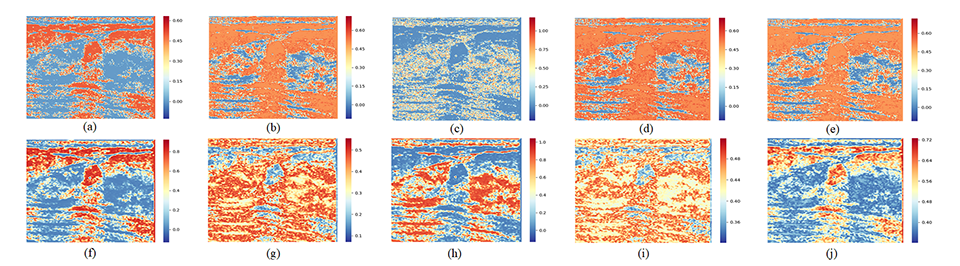}}
    \caption{Heatmaps of the uncertainty maps: (a)-(e) are the uncertainty maps of tumor, fat layer, mammary layer, muscle layer, and background, which are generated by the trainable Sigmoid membership function; (f)-(j) are generated by the trainable Gaussian membership function. Blue represents low uncertainties and red represents high uncertainties}
\end{figure*}
Heatmaps in Fig. 8 show the uncertainties in different categories. The red areas have high uncertainties, and blue areas have low uncertainties in corresponding categories. To compute the overall categories uncertainty, a fuzzy AND operation is applied as shown in Eq. (8).
\begin{equation}
    o_i = min_r x_i^r, \, r=0,1,2,3,4
\end{equation}
The overall categories uncertainty maps on gray-level intensities are shown as the heatmaps in Fig. 9.\par
\begin{figure}[tbp]
    \centerline{\includegraphics[width=8.5cm,height=4.5cm]{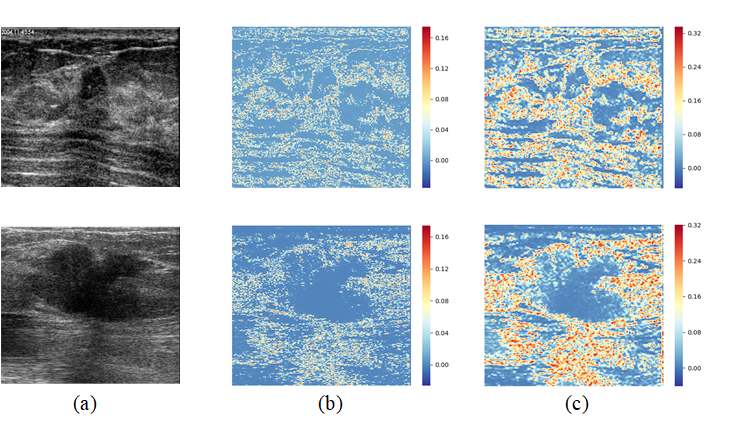}}
    \caption{Heatmaps of overall categories uncertainty maps: (a) original images; (b) overall categories uncertainty maps generated by using the trainable Sigmoid membership function; (c) overall categories uncertainty maps generated by using the trainable Gaussian membership function. Blue represents low uncertainties and red represents high uncertainties}
\end{figure}
From Fig. 9, it can be observed that the pixels on the boundaries between categories have high uncertainties. The pixels in mammary layer and muscle layer have high uncertainties as well. Trainable Sigmoid and Gaussian membership functions can obtain similar results. 

\textbf{Fusion of original convolutional network and fuzzy information:} To reduce the uncertainty on the original channel, the overall categories uncertainty maps are fused with the corresponding original channels as shown in Eq. (9). 
\begin{equation}
    o_i = (1-x_i) \cdot I_i
\end{equation}
where $x_i$ are the overall categories uncertainty maps obtained by Eq. (8), and $I_i$ are the original channels of the input. It means if a pixel has high uncertainty, its weight should be reduced.

The inputs of the network are three channel images \cite{ref16} (shown in Fig. 3 (a)). The first channel is the original image after contrast enhancement; the second channel is the wavelet low frequency information; and the third channel is the wavelet high frequency information. The fuzzification layer, uncertainty representation layer, and fusion layer are applied to all three channels separately. The results after reducing uncertainty are shown in Fig. 10. The boundary areas in (c) and (d) are more distinct than that in (a).\par
\begin{figure}[tbp]
    \centerline{\includegraphics[width=8.5cm,height=3.5cm]{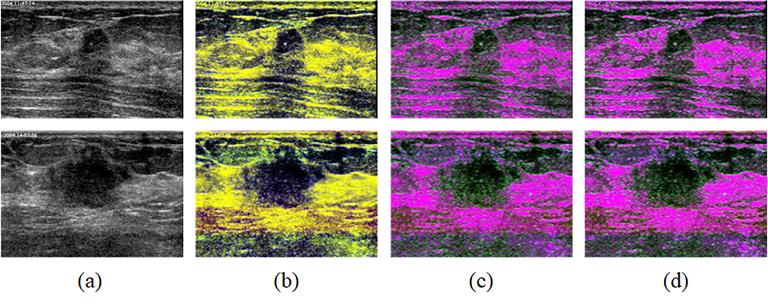}}
    \caption{The fusion of uncertainty maps and input images: (a) original images; (b) 3-channel images with gray-level intensity and wavelet information; (c) resulted images using the Sigmoid function and Eq. (9); (d) resulted images using the Gaussian function and Eq. (9).}
\end{figure}
The resulted image inputs to the convolutional layer for obtaining the convolutional feature maps. The network structure is similar to U-Net. The first convolutional layer has a feature map of 64 dimensions. All the 64-dimensional features will conduct fuzzification, uncertainty representation, and fusion with original information.
\subsection{Fuzzy information guided fully convolutional network training}
The uncertainty maps multiply with the input images and the results input to the first convolutional layer. The entire network structure is shown in Fig. 11. \par
\begin{figure*}[tbp]
    \centerline{\includegraphics[width=18cm,height=3cm]{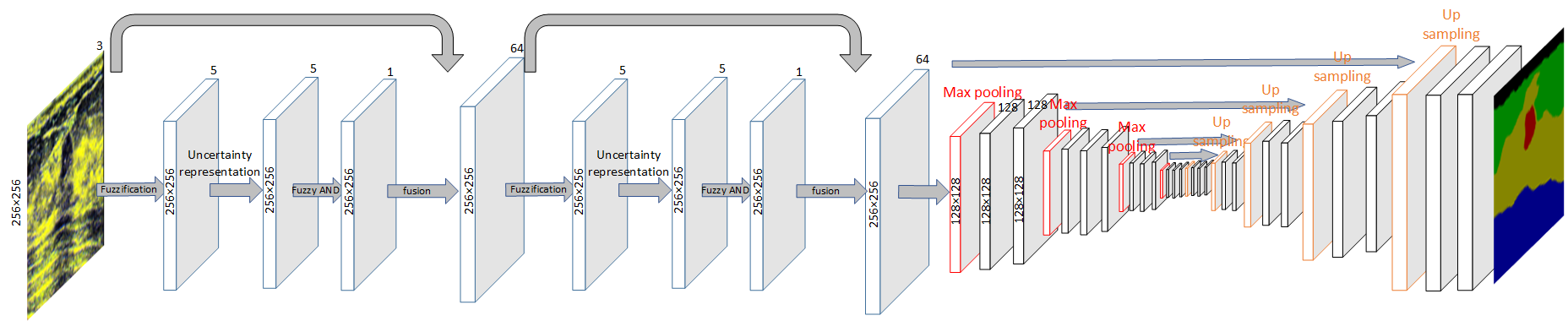}}
    \caption{Structure of the proposed fuzzy FCN.}
\end{figure*}
The output is processed by pixel-wised soft-max which is defined as \cite{ref14}:
\begin{equation}
    p(x) = exp(a(x)) / \sum_{r=0}^4 {exp(a_r(x))}
\end{equation}
where $a(x)$ is the output of the neural network, $r$ represents the class index, and $x$ represents the input pixel. The cross-entropy loss function is computed by the output probability and the label of each pixel:
\begin{equation}
    C = - \sum_{x\in X} {q(x)\log{(p(x))}}
\end{equation}
where $q(x)$ is the pixel label which is background, tumor area, fat layer, mammary layer or muscle layer with one-hot encoding. The original parameters in U-Net are initialized by uniform distribution. If using trainable Sigmoid membership function, parameter $b_i^r$ in Eq. (4) is initialized by the mean of all training samples in category \textit{r}. Parameter $a_i^r$ in Eq. (4) is initialized by uniform distribution. The parameters $\mu_i^r$ and ${\sigma_i^r}^2$ in Eq. (5) are initialized by the mean and variance of the intensities of all training samples in category \textit{r}. The training strategy is based on the back-propagation algorithm. All of the functions should be differentiable. Functions in fuzzy layer (trainable Sigmoid function or trainable Gaussian function) are all differentiable. The training strategy is in Algorithm 1.
\renewcommand{\algorithmicrequire}{\textbf{Input:}}  
\renewcommand{\algorithmicensure}{\textbf{Initialization:}} 

\begin{algorithm}[htbp]
  \caption{Training Strategy of Fuzzy Fully Convolutional Network}
  \label{alg::conjugateGradient}
  \begin{algorithmic}[1]
    \REQUIRE
      \textit{M} training images: each is resized to 256 $\times$ 256; pixel-wise labels of the \textit{M} samples; category number \textit{r}; input channel number D; training rate $\eta$, training epoch number \textit{S}; batch size \textit{P}; and training decrease rate $\epsilon$.
    \ENSURE
      Fuzzy transform has $2 \times r \times n \times {\rm D}$ parameters, where $n$ is the number of pixels. Parameters in fuzzy layer use the mean and variance of the training samples in each category to initialize. Other parameters are initialized by uniform distribution.
    \FOR {$t = 1, 2, …, S$} 
        \FOR {$m = 1, 2, …, \frac{M}{P}$} 
            \STATE
            Input a batch of images to the network and obtain the error of loss function in Eq. (11).
            \STATE Compute the weight changing rate $\nabla\omega$ using the back-propagation algorithm and the error of loss function for all the parameters in fuzzy layer and original U-Net. Then, compute the new parameters using Adam method and the learning rate $\eta$.  
        \ENDFOR
        \STATE Update the learning rate by the decrease rate $\epsilon$.
    \ENDFOR
    \renewcommand{\algorithmicensure}{\textbf{Output:}}
    \ENSURE
      Weight vector of the neural network
  \end{algorithmic}
\end{algorithm}
The fuzzy fully convolutional network deals with the following issues: 1) it can reduce the uncertainty, and 2) it can solve small sample size problem, and it can even replace the information extension process \cite{ref16} (experimental details will be discussed in Section 4).

\section{Breast-anatomy constrained post-processing}
The fuzzy information guided fully convolutional network (FCN) can perform BUS image segmentation. However, the segmentation results are not good because the dataset size is too small, and the network structure is quite deep. Fully connected conditional random fields (CRFs) are often utilized to refine the segmentation results. In \cite{ref28}, Krahenbuhl, et al. provided an approximation algorithm to fully connected CRFs for multi-objects segmentation. The approximation algorithm increases the efficiency of fully connected CRFs and makes it possible for semantic segmentation. Chen et al. proposed the Deeplab structure for nature image semantic segmentation by applying the atrous convolutional operation and atrous spatial pyramid pooling (ASPP) \cite{ref26,ref27}. In addition, Deeplab was also applied fully connected CRFs to the end of the architecture for achieving better performance. Zheng et al. realized CRFs using a recurrent neural network (RNN) \cite{ref22}. It made the FCN+CRFs structure become a deep end-to-end architecture. Fully connected CRFs took care of the relationships among pixels, not just classified pixels into different categories. The physical location of a pixel and the features can affect the final segmentation. 

In order to involve the label context, Liu et al. provided a Markov random field (MRF) method with the mixture of label context \cite{ref19}. The MRF was realized by Deep Parsing Network (DPN). In \cite{ref16}, three categories: tumor, mammary layer, and background were classified. The importance of the locations of mammary layer and tumor was discussed, since most of the breast cancers begin from the cells in the mammary layer \cite{ref20}. 

In this paper, the proposed method will be applied to five categories. Which means that not only mammary layer and tumor are determined, but other layers such as muscle layer, fat layer, and background are also involved. 
\subsection{Fully Connected CRFs}
In image segmentation, the pixels are considered as a random field ${\rm I}=\{I_1,I_2,…,I_N\}$, where $I_N$ represents the features of the \textit{N}th pixel, and the labels are considered as another random field ${\rm X}=\{x_1,x_2,…,x_N\}$, where $x_N$ represents the label of the \textit{N}th pixel. The segmentation task is formulized by CRFs through computing the conditional probability $P({\rm X|I})$ which is characterized by Gibbs distribution \cite{ref28}:\par
\begin{equation}
    P({\rm X|I}) = \frac{1}{Z({\rm I})}exp(-\sum_{w \in W}{\phi_w({\rm X}_w|{\rm I})})
\end{equation}
where \textit{W} is a graph on random field X, and \textit{w} represents cliques in the graph. In the fully connected CRFs, \textit{W} is a complete graph on X, and a pixel has edges connected to all other pixels in the image. $\sum_{w \in W}{\phi_w({\rm X}_w|{\rm I})}$ in Eq. (12) is the Gibbs energy function \cite{ref28}:\par
\begin{equation}
\begin{split}
    &E({\rm X|I})=\sum_{w \in W}{\phi_w({\rm X}_w|{\rm I})}\\
    &E({\rm X}) = \sum_i{\theta_i(x_i)} + \sum_i\sum_j{\theta_{ij}(x_i, x_j)}
\end{split}
\end{equation}

Segmentation could also be treated as the optimum labeling problem which modeled the maximum a posteriori problem (MAP). The maximum probability $P({\rm X|I})$ is the same as the minimum energy function $E({\rm X})$. The first term in Eq. (13), $\theta_i(x_i)=-log{P(x_i)}$ is a unary potential function, which is provided by a unary classifier, such as FCN. In the proposed approach, the probability is computed by the neural network in Section 2. $\theta_{ij}(x_i,x_j)$ is the pairwise potential function \cite{ref28}:
\begin{equation}
    \theta_{ij}(x_i,x_j) = \mu(x_i,x_j)\sum_m{\omega_mk_m(f_i,f_j)}
\end{equation}
where $\mu(x_i,x_j)=1$, if $x_i \neq x_j$, and $\mu(x_i,x_j)=0$ if $x_i=x_j$, which is known as the Potts model. This coefficient shows if two pixels are in the same category, the energy is minimum. $k_m$ is a Gaussian kernel, where $f_i$ and $f_j$ are the features of pixels $i$ and $j$. $\omega_m$ is the combination weight of the \textit{m}th Gaussian kernel. There are two Gaussian kernels in \cite{ref28}. In the first Gaussian kernel, the feature is defined on the physical position of the pixels, and another is the color feature of the pixels. In this research, the color feature (RGB) represents the gray-level information in R channel, approximation coefficient of wavelet transforms in G channel, and high frequency information of wavelet transforms in B channel. If the input image is not preprocessed, only intensity combined with position is used. The second Gaussian kernel is only defined on the positions of pixels. The detail of the pairwise potential function is shown in Eq. (15) \cite{ref28}:
\begin{equation}
\begin{split}
    \sum_m{\omega_mk_m(f_i,f_j)}=&\omega_1exp(-\frac{\|p_i-p_j\|^2}{2\sigma_\alpha^2}-\frac{\|I_i-I_j\|^2}{2\sigma_\beta^2})\\
    &+\omega_2exp(-\frac{\|p_i-p_j\|^2}{2\sigma_\gamma^2}),m=1,2
\end{split}
\end{equation}
where $p_i$ represents the position of the \textit{i}th pixel, and $I_i$ represents the color feature of the \textit{i}th pixel. $\sigma_\alpha^2$, $\sigma_\beta^2$, and $\sigma_\gamma^2$ are the parameters of CRFs determined by experiments. The first term $\omega_1k_1(f_i,f_j)$ is the appearance kernel. It encourages two pixels with the similar color features ($I_i$) and close positions ($p_i$) to be in the same category; otherwise, classifies them into different categories. The second term is the smooth kernel which only depends on pixel position. It helps to smooth the segmentation result. 
\subsection{Medical knowledge in BUS images}
The fully connected CRFs in Section 3.1 are popular for nature image segmentation. In this research, the target image is BUS image, which contains special regular patterns. As shown in Fig. 1 (b) the BUS images have the following properties:
1) The anatomy of human breast consists of skin (SK), subcutaneous fat (SF), intraglandular fat (IF), glandular tissue, retromammary fat (RF), and muscle \cite{ref39}. We can simplify the breast model as the layer structure shown in Fig. 1(b). On the top is the skin layer. The subcutaneous fat layer is beneath the skin layer. The mammary layer is below the fat layer and followed by the muscle layer.
2) Breast cancer is usually ellipse-shaped and begins from the cells in mammary layer. In most cases, breast cancer stays inside the mammary layer.

In this study, the skin layer is treated as background because the number of samples containing skin layer is quite small. However, due to the position, the skin layer is different from the retro-muscle background area. In order to make the context of different layers more reasonable, the skin layer is treated as pre-fat background area. In general, the contexts of pre-fat background area, fat layer, mammary layer, muscle layer, retro-muscle layer, and breast tumor are used. $V_i$ represents the category of pixel \textit{i} assigned by FCN, $V_i \in \{L^1,L^2,L^3,L^4,L^5,L^6\}$. $L^1$, $L^2$, $L^3$, $L^4$, $L^5$, and $L^6$ represent pre-fat background area, fat layer, mammary layer, muscle layer, retro-muscle layer, and tumor, respectively (Fig. 12 (b)). In this research, they are three-dimensional vectors.
\begin{figure}[tbp]
    \centering
	\subfloat[]{\includegraphics[width=4cm,height=3cm]{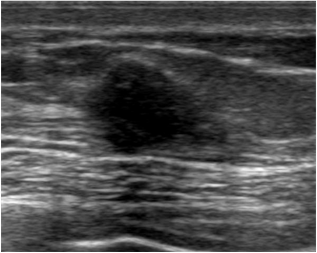}}\quad
	\subfloat[]{\includegraphics[width=4cm,height=3cm]{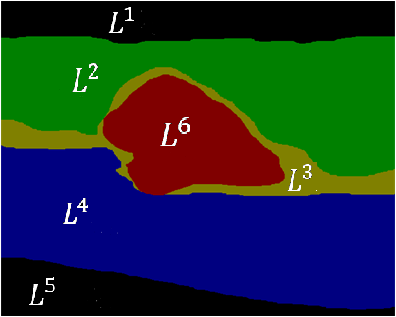}}\\
	\captionsetup{justification=centering}
	\caption{Breast anatomy: (a) BUS image; (b) breast anatomy obtained by FCN.}
\end{figure}
\subsection{Breast-anatomy constrained fully connected CRFs}
As discussed in Section 3.2, the breast cancer usually begins in the mammary layer. However, some pixels in the fat layer and muscle layer might be classified into wrong categories; in addition, the pixels in muscle layer have similar intensity levels to that of the pixels in mammary layer which may also cause misclassification. The medical knowledge can overcome this problem. After locating the positions of the fat layer, mammary layer and muscle layer, the context information can be used to prevent the wrongly classified patches in each layer. The original fully connected CRFs contain the energy function in Eq. (14)-(15). The Gaussian kernel in Eq. (15) consists of pixel positions and color features. To involve breast anatomy, the category of pixel \textit{i} assigned by FCN, which is defined as $V_i$, is treated as another feature. A new Gaussian kernel based on $V_i$ and position of the pixel ($p_i$) is utilized. The new energy function contains three terms:
\begin{equation}
\begin{split}
    &\sum_m{\omega_mk_m(f_i,f_j)}=\omega_1exp(-\frac{\|p_i-p_j\|^2}{2\sigma_\alpha^2}-\frac{\|I_i-I_j\|^2}{2\sigma_\beta^2})\\
    &+\omega_2exp(-\frac{\|p_i-p_j\|^2}{2\sigma_\gamma^2})\\
    &+\omega_3exp(-\frac{\|p_i-p_j\|^2}{2\sigma_\tau^2}-\frac{\|V_i-V_j\|^2}{2\sigma_\lambda^2}),m=1,2,3
\end{split}
\end{equation}
where $exp(-\frac{\|p_i-p_j\|^2}{2\sigma_\tau^2}-\frac{\|V_i-V_j\|^2}{2\sigma_\lambda^2})$ is the Gaussian kernel of the context information, and $V_i$ and $V_j$ represent categories of pixel \textit{i} and \textit{j} assigned by the fully convolutional neural network in Section 2. $\sigma_\tau^2$ and $\sigma_\lambda^2$ are the parameters of CRFs. 

Here, two context distances are defined: 1) context distance between two pixels, $\|V_i-V_j\|^2$ where $i$ and $j$ represent pixel $i$ and $j$, and 2) context distance between two categories, $\|L^s-L^t\|^2$ where $s$ and $t$ represent the category indexes. In this research, $1 \leq s$, $t \leq 6$, $i \in \mathbf{Z}$. $\|L^s-L^t\|^2$ is the Euclidean distance of the two category vectors. $\|V_i-V_j\|^2$ is the context distance between category of pixel $i$ and category of pixel $j$. For example, if the pixel $i$ is in category $L^1$, and pixel $j$ is in category $L^2$, $\|V_i-V_j\|^2$ equals to $\|L^1-L^2\|^2$.\par
\begin{figure}[tbp]
    \centering
	\subfloat[]{\includegraphics[width=4cm,height=4cm]{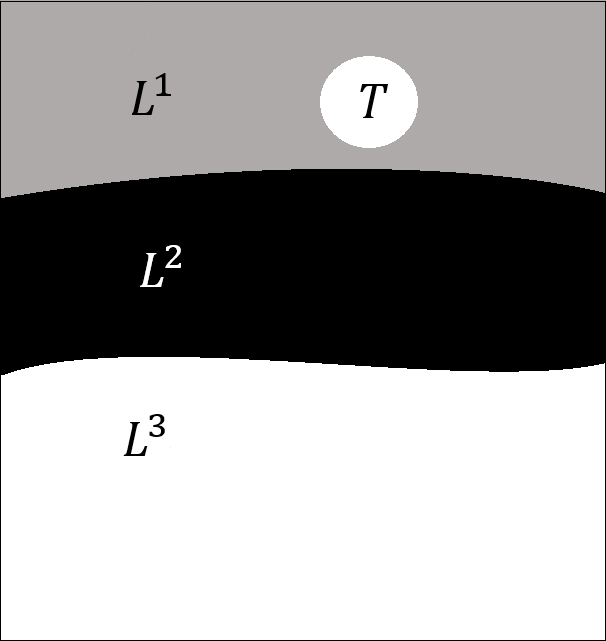}}\quad
	\subfloat[]{\includegraphics[width=4cm,height=4cm]{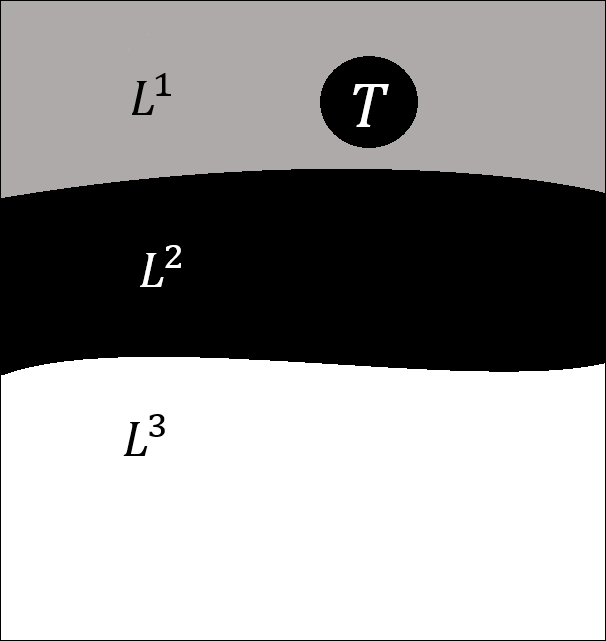}}\\
	\captionsetup{justification=centering}
	\caption{Simulated image to show the context distance among categories}
\end{figure}
To demonstrate how to utilize the context distance between pixels and categories and its effectiveness on BUS image segmentation, simulated images are utilized shown in Fig. 13. In Fig. 13 (a), $L^1$, $L^2$ and $L^3$ represent three categories. $T$ represents a wrongly classified patch, which should be in $L^1$, but assigned to $L^3$ by the unary classifier. If the context distances among three categories are set as: the context distance between $L^1$ and $L^2$ equals to the context distance between $L^2$ and $L^3$; the context distance between $L^1$ and $L^2$ is greater than the context distance between $L^1$ and $L^3$; then pixels in $T$ has the chance to be corrected into $L^1$. Here, four pixels are chosen to demonstrate how it works: 1) pixel $i$ in area $T$; 2) pixel $h$ in area $L^1$; 3) pixel $v$ in area $L^2$; 4) pixel $j$ in area $L^3$. Pixel $i$ is in area $T$ and area $T$ is now in category $L^3$; pixel $v$ is in area $L^2$, so the context distance between pixel $i$ and pixel $v$ equals to the context distance between categories $L^3$ and $L^2$ as introduced in the previous paragraph, i.e. $\|V_i-V_v\|^2=\|L^3-L^2\|^2$. For other pixels, the situations are the same, i.e. $\|V_i-V_h\|^2=\|L^3-L^1\|^2$; $\|V_i-V_j\|^2=\|L^3-L^3\|^2=0$. Therefore, $\|V_i-V_v\|^2=\|L^3-L^2\|^2>\|V_i-V_h\|^2=\|L^3-L^1\|^2>\|V_i-V_j\|^2=\|L^3-L^3 \|^2$ because of the assumption made before. Meanwhile, $\|p_i-p_j\|^2>\|p_i-p_v \|^2>\|p_i-p_h\|^2$ because of the physical positions of these pixels, where $p_i$, $p_h$, $p_v$ and $p_j$ are the positions of these pixels in Eq. (16). Hence, the pixels in area $T$ have smaller context distances with pixels in $L^1$ than that in $L^2$; and the pixels in area $T$ have smaller space distances with the pixels in $L^1$ than that in $L^2$, which means pixels in area $T$ could not be in category $L^2$. Even if the pixels in area $T$ have zero context distances with the pixels in $L^3$ (i.e. they are in the same category), they have smaller space distances with the pixels in $L^1$ than that in $L^3$. Therefore, the pixels in area $T$ still have the possibility to be classified into category $L^1$. In Fig. 13 (b), the pixels in area $T$ are wrongly classified into category $L^2$ and the pixels in area $T$ have the same context distances with the pixels in $L^1$ and $L^3$ because of the assumption before, but they have smaller space distances with the pixels in $L^1$ than that in $L^3$. Even if the pixels in area $T$ have zero context distances with the pixels in $L^2$, their space distances with the pixels in $L^1$ are smaller than that with the pixels in $L^2$. Therefore, the pixels in area $T$ still have the chance to be classified into $L^1$ by properly setting weight $\omega_3$ and parameters $\sigma_\tau^2$ and $\sigma_\lambda^2$ in Eq. (16). The BUS images (Fig. 12 (a) and (b)) are similar to the simulated examples.

The context distances between the categories can be classified into three classes (Fig. 12 (b)) in the BUS images: 1) two layers are neighbors to each other ($D_1$), e.g. fat layer ($L^2$) and mammary layer ($L^3$); 2) two layers are separated by another layer ($D_2$), such as fat layer ($L^2$) and muscle layer ($L^4$); 3) two layers are separated by two layers ($D_3$), such as fat layer ($L^2$) and retro-muscle background area ($L^5$).
\begin{equation}
\begin{split}
    D_1&=\|L^i-L^{i+1}\|^2,1 \leq i \leq 4, i \in \mathbf{Z}\\
    D_2&=\|L^i-L^{i+2}\|^2,1 \leq i \leq 3, i \in \mathbf{Z}\\
    D_3&=\|L^i-L^{i+3}\|^2,1 \leq i \leq 2, i \in \mathbf{Z}
\end{split}
\end{equation}

The relations among them are:\par
\begin{equation}
D_1 > D_2 > D_3
\end{equation}

The reason of setting such relations among them (Eq. (18)) is to follow the situation in the simulated example. The relations encourage a clear boundary and void wrongly classified patches like $T$ in Fig. 13. $L^1$ and $L^5$ are both treated as the background in FCN. Here, they are treated as different labels for easier description. They have high space distance. Their space distance plays more important role than context distance in this term, so their context distance is not involved. After defining the context distances among five layers, the context distances between tumor and five layers could be defined. The tumor ($L^6$) usually locates in the mammary layer ($L^3$). Sometimes, the mammary layer above the tumor or below the tumor is very thin; and the tumor seems to be in the fat layer ($L^2$) or muscle layer ($L^4$). The context distance between tumor ($L^6$) and mammary layer ($L^3$) should be the largest, because it encourages a clear boundary between tumor and mammary layer. The context distances between the tumor ($L^6$) and fat layer ($L^2$) or muscle layer ($L^4$) should be the second largest. This gives the chance to correct some wrongly classified patches in these layers. The context distance between tumor ($L^6$) and the background ($L^1$ and $L^5$) should be the smallest. Because some background areas are likely classified as the tumors, and such situation should be voided (refer Fig. 13). The relationships are shown in Eq. (19).\par
\begin{equation}
\begin{split}
\|L^6-L^3\|^2>&\|L^6-L^2\|^2 \approx\|L^6-L^4\|^2\\
>&\|L^6-L^1\|^2 \approx \|L^6-L^5\|^2
\end{split}
\end{equation}

The category vectors $L^1$, $L^2$, $L^3$, $L^4$, $L^5$, and $L^6$ should satisfy the constraints in Eq. (17)-(19) to realize the medical anatomy constraints. By solving Eq. (17)-(19), $L^1=\{61.2, 20, 15\}$, $L^2=\{25, 37.1, 0\}$, $L^3=\{40, 0, 0\}$, $L^4=\{55, 37.1, 0\}$, $L^5=\{18.8, 20.7, 15\}$, and $L^6=\{40, 30, 26.5\}$. For $D_1 \approx 40$, $D_2 \approx 30$, $D_3 \approx 23$, $\|L^6-L^3\|^2=40$, $\|L^6-L^2\|^2 \approx \|L^6-L^4\|^2 \approx 30$, $\|L^6-L^1\|^2 \approx \|L^6-L^5\|^2 \approx 26$. The relations among context labels are shown in Fig. 14. If a pixel is classified into category $L^s, s=1, 2, 3, 4, 5, 6$, a category map will be created and the corresponding pixel in the category map will be assigned by the value of $L^s$. The category map is used as another feature in Eq. (16).\par
\begin{figure}[tbp]
    \centerline{\includegraphics[width=5cm,height=4cm]{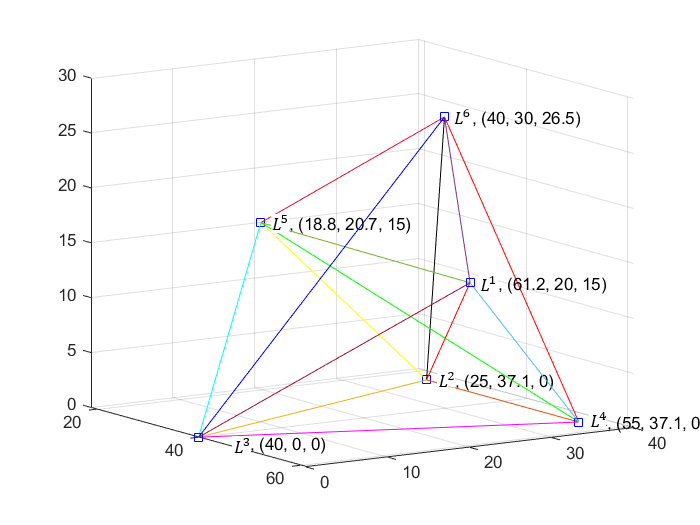}}
    \caption{The coordinates of the labels}
\end{figure}

By setting the label vectors with these values, the proposed CRFs energy function encourages two pixels whose space distances and context distances are both small to be in the same category. It will remove some wrongly classified patches.

The mean field approximate algorithm \cite{ref28} is used to solve the fully connected CRFs energy minimization problem, which uses the fast algorithm of the high-dimensional Gaussian filter \cite{ref29} to reduce the time complexity. 
\section{Experimental results}
\subsection{Dataset}
The performances of the proposed fuzzy FCN and the breast-anatomy constrained fully connected CRFs are evaluated by a dataset of 325 images. Image 1 to 141 were collected over 10 years by the Second Affiliated Hospital of Harbin Medical University using VIVID 7 (GE) and EUB-6500 (Hitachi) imaging systems. Image 142 to 325 were collected in recent 3 years by the First Affiliated Hospital of Harbin Medical University using Aixplorer Ultrasound system (SuperSonic Imagine). The resolution of the first 141 images is 550 $\times$ 450, and the rest 184 images have the resolution 787 $\times$ 526. Informed consents to the protocol from all patients were acquired. The privacy of the patients is well protected.

An experienced radiologist from the First Affiliated Hospital of Harbin Medical University delineated the boundaries of the layers and tumors. The pixel-wise ground truths are generated according to the manually delineated boundaries. The experiment is conducted in two steps. In the first step, fuzzy FCN is applied and compared with U-Net \cite{ref14}, FCN \cite{ref21}, and \cite{ref16}. In the second step, the breast-anatomy constrained fully connected CRFs are applied. Then, the overall performance of the proposed method is computed. The final tumor segmentation results will be compared with that of the existing methods \cite{ref13,ref16,ref31,ref32,ref33,ref34,ref35,ref37}.
\subsection{Evaluation metrics}
Three area metrics are used to evaluate the performance: true positive rate (TPR), false positive rate (FPR), and intersection over union (IoU) \cite{ref36,ref37}. The IoU for every category are computed, and the mean over 5 categories IoUs is used as the overall performance. The TPR and FPR for tumors are used to compare with that of the previous tumor segmentation methods. Due to the limitation of the number of the samples, 10-fold validation is used. The samples are randomly divided into 10 subsets. Each time, 9 of them are used for training and 1 is used for testing. The metrics are computed by Eq. (20):
\begin{equation}
\begin{split}
{\rm TPR}=&|A_r \cap A_m|/|A_m|\\
{\rm FPR}=&|A_r \cup A_m - A_m|/|A_m|\\
{\rm IoU}=&|A_r \cap A_m|/|A_r \cup A_m|
\end{split}
\end{equation}
where $A_r$ is the region generated by the proposed method or existing methods, and $A_m$ is the region of the ground truth.
\subsection{Segmentation result of fuzzy FCN}
In order to show the effectiveness of the fuzzy logic, the proposed fuzzy layer is applied to U-Net. The feature extension method \cite{ref16} is also employed for comparison. Histogram equalization is applied to the input images, then the wavelet transform is utilized. Five networks are trained: 1) the gray-level images (original images) are used to train the U-Net; 2) the image after preprocessing and wavelet transform are used to train the U-Net \cite{ref16}; 3) the fuzzification layer is applied, and the output of the uncertainty representation layer is combined with the input image. The fuzzy layer is also applied to the feature maps of the first convolutional layer; 4) to demonstrate the existence of the uncertainty in BUS images and the effectiveness of the fuzzy layer; wavelet transform is removed, and only the gray-level BUS image is used as the input; and 5) the FCN with VGG16 network structure using original gray-level image as input and pretrained by nature images.\par
\begin{figure*}[tbp]
    \centerline{\includegraphics[width=18cm,height=9.5cm]{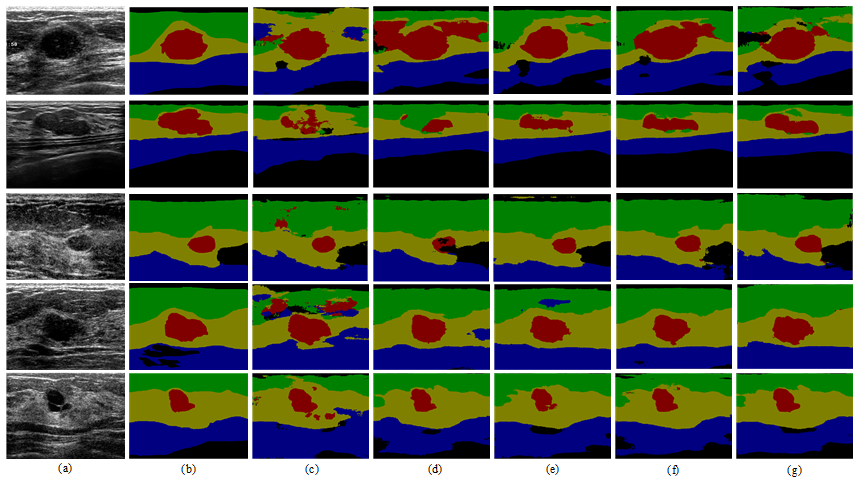}}
    \caption{The semantic segmentation results. (a) the original images; (b) the ground truths; (c) results of U-Net with gray-level images as the inputs; (d) results of U-Net with information extension [16]; (e) results of the proposed fuzzy FCN with trainable Gaussian membership function and 3-channel image; (f) results of the proposed fuzzy FCN with trainable Sigmoid membership function and 3-channel image; (g) results of the proposed fuzzy layer with trainable Sigmoid membership function and gray-level image.}
\end{figure*}

Fig. 15 (b) shows the pixel-wise ground truths. The black area is background; the green area is the fat layer; the yellow area is the mammary layer; the blue area is the muscle layer; and the red area is the tumor. Fig. 15 (c) shows the results of the U-Net with original gray-level images as the inputs, which are the worst. Using information extension method \cite{ref16} can improve the performance in some cases (Fig. 15 (c)); however, sometimes adding wavelet information can make the results worse. If adding fuzzy processing and reducing uncertainty in the 3-channel input images, the results are better (Fig. 15 (e) and (f)). Even if not applying wavelet transform and preprocessing, the fuzzy FCN can still achieve good results. In Table I, the evaluation of the 6 networks is listed. The proposed methods can improve BUS image semantic segmentation. The IoU on tumor is 78.53\% by using fuzzy FCN with 3-channel image and trainable Sigmoid membership function. It achieves a 4\% improvement than that of non-fuzzy FCN. The overall IoU over the 5 categories is 78.32\% using fuzzy FCN with 3-channel image and trainable Sigmoid membership function and has a 0.7\% improvement than that of the non-fuzzy FCN.\par
\begin{table*}[htbp]
\begin{center}
\caption{Evaluation results on 325 cases dataset. Evaluation metric is IoU}
\begin{tabular}{p{5cm}|ccccccccc}
\hline
 & Fat & Mammary & Muscle & Background & Tumor & Mean \\\hline
U-Net \cite{ref14} with original image  & 70.34 & 66.72	& 66.17	& 65.91	& 74.66 & 68.76 \\
U-Net with 3-channel image \cite{ref15}	& 84.05	& 75.92	& 74.89	& 78.35	& 74.88 &	77.62 \\
FCN-VGG16 \cite{ref16} with original image using pretrained model & 82.57 & 75.47 &	75.53 & \textbf{78.59} & 74.42 & 77.32 \\\hline
Fuzzy FCN with 3-channel image and Sigmoid membership function	& \textbf{84.07} & 76.01 & 74.62 & 78.39 & \textbf{78.53} & \textbf{78.32} \\
Fuzzy FCN with 3-channel image and Gaussian membership function	& 83.47	& 74.73 & 73.95 & 77.51 & 75.32 & 70.00 \\
Fuzzy FCN with original image and Sigmoid membership function & 82.56 &	\textbf{76.14} & \textbf{74.64} & 75.98 & 77.56 & 77.38 \\\hline
\end{tabular}
\label{tab1}
\end{center}
\end{table*}

Moreover, if not using the information extension method (just using the original images as inputs without using preprocessing and wavelet transform), the fuzzy FCN achieves (the bottom row in Table I) a 3\% improvement on tumor than the non-fuzzy FCN with gray-level image. The overall IoU is close to that of the non-fuzzy FCN with 3-channel image and achieves a 9.6\% improvement than the non-fuzzy FCN with gray-level image.  
\subsection{Breast-anatomy constrained fully connected CRFs.}
Breast anatomy constrained fully connected CRFs utilize the medical context information. The original fully connected CRFs and the approximation algorithm in \cite{ref28} are employed. It has three effects: 1) correct the wrongly classified pixels; 2) make the boundaries between layers more accurate; and 3) increase the overall segmentation performance. The CRFs parameters $\omega_m$, $\sigma_\alpha^2$, $\sigma_\beta^2$, $\sigma_\tau^2$, and $\sigma_\lambda^2$ are determined by experiments, and the medical context label and the context distance relation are shown in Fig. 14. The segmentation results are shown in Fig. 16. In Table II, the output of Fuzzy FCN with 3-channel image and trainable Sigmoid membership function is used as the unary energy in CRFs model. The original CRFs and proposed CRFs are for comparison. \par
\begin{table*}[htbp]
\begin{center}
\caption{Evaluation results of breast anatomy constrained CRFs. Evaluation metrics using IoU}
\begin{tabular}{p{4cm}|ccccccccc}
\hline
 & Fat & Mammary & Muscle & Background & Tumor & Mean \\\hline
Fuzzy FCN with 3-channel image and Sigmoid membership function + CRFs &	81.52 & \textbf{78.63} & 75.24 & 76.48 & 79.32 & 78.24 \\\hline
Fuzzy FCN with 3-channel image and Sigmoid membership function & 84.07 &	76.01 & 74.62 & 78.39 & 78.53 & 78.32\\
Fuzzy FCN with 3-channel image and Sigmoid membership function + Proposed CRFS & \textbf{85.06} & 77.24 & \textbf{78.66} & \textbf{80.09} & \textbf{81.29} & \textbf{80.47} \\\hline
\end{tabular}
\label{tab2}
\end{center}
\end{table*}
\begin{figure*}[tbp]
    \centerline{\includegraphics[width=14cm,height=9.5cm]{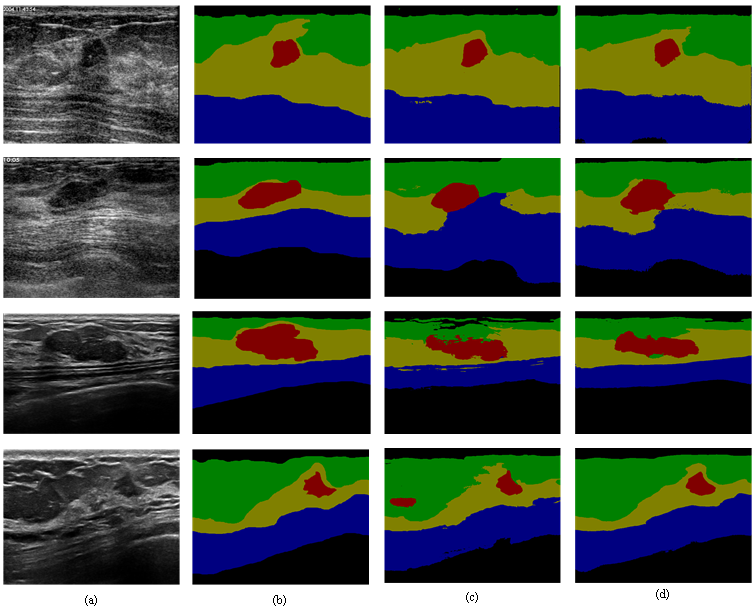}}
    \caption{Segmentation results of breast-anatomy constrained fully connected CRFs: (a) original images; (b) ground truths; (c) the results of fully connected CRFs; (d) the results of the proposed method.}
\end{figure*}

In Fig. 16 (c), the original CRFs classify some pixels wrongly. For example, in the fourth row, there are pixels in the fat layer classified into the tumor. The proposed CRFs utilize the medical context constraints to overcome such problem. The same as in the second row, the muscle layer grows into the mammary layer using the original CRFs, and in the third row, background area and fat layer interlace each other. 

Table II shows the IoU of each category and the overall mean IoU. The proposed method achieves 81.29\% of IoU for tumor, and 80.47\% of overall IoU. In the results of both tumor and overall IoU, the proposed method achieves about 2\% improvements than that of the original CRFs.
\subsection{Tumor segmentation results and comparison with existing segmentation methods}
The existing methods only focus on breast cancer segmentation, while semantic segmentation methods work on multi-object segmentation. In this section, the proposed method and the methods in \cite{ref9,ref31,ref33,ref34,ref35} are compared. The semi-automatic BUS image segmentation methods \cite{ref9,ref31} were studied, in which the regions of interest (ROIs) were given, and the methods could segment the tumor areas automatically. The fully automatic BUS image segmentation methods were studied \cite{ref33,ref34,ref35}. The tumor segmentation results are shown in Fig. 17.

In Fig. 17, the semi-automatic segmentation methods (Fig. 17(c) and (d)) obtain good results. Semi-automatic segmentation methods are useful when doctors focus on specific areas and operate with the CAD systems interactively. Existing fully automatic segmentation methods obtain worse results, since the performance of these methods relied on the individual dataset. They can obtain good performance only using own datasets and need huge number of training samples. The proposed method can achieve the best result even on small dataset, and its robustness is much higher than that of other methods in comparison.\par
\begin{figure*}[tbp]
    \centerline{\includegraphics[width=17cm,height=5cm]{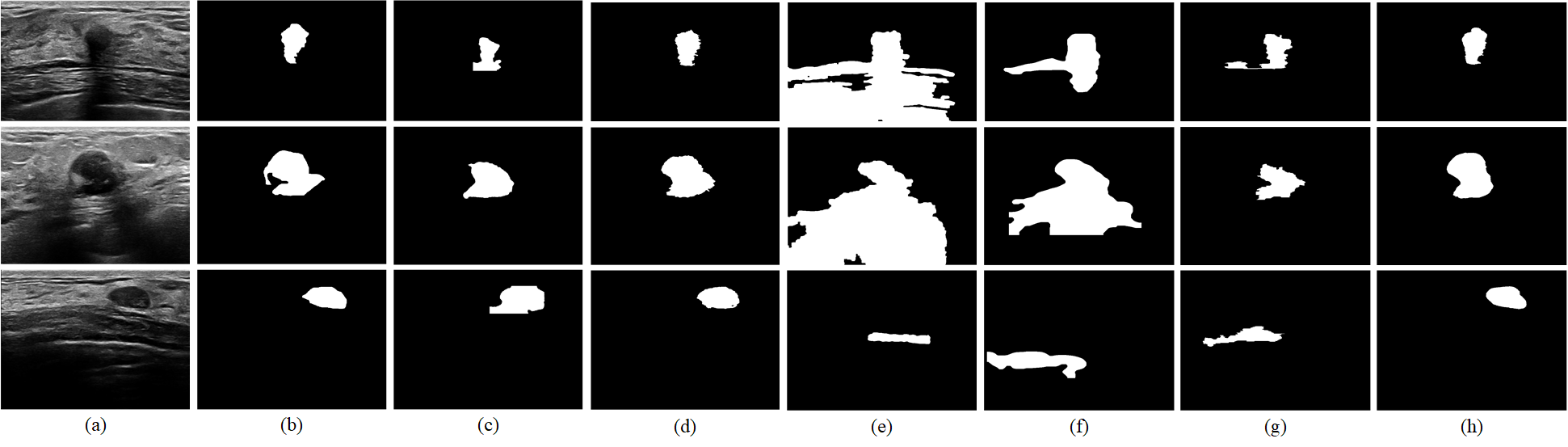}}
    \caption{Tumor segmentation results of the proposed method and existing methods: (a) the original images; (b) ground truths; (c) results using \cite{ref9}; (d) results using \cite{ref31}; (e) results using \cite{ref34}; (f) results using \cite{ref33}; (g) results using \cite{ref35}; (h) results using the proposed method.}
\end{figure*}
\begin{figure*}[tbp]
    \centerline{\includegraphics[width=12cm,height=3.5cm]{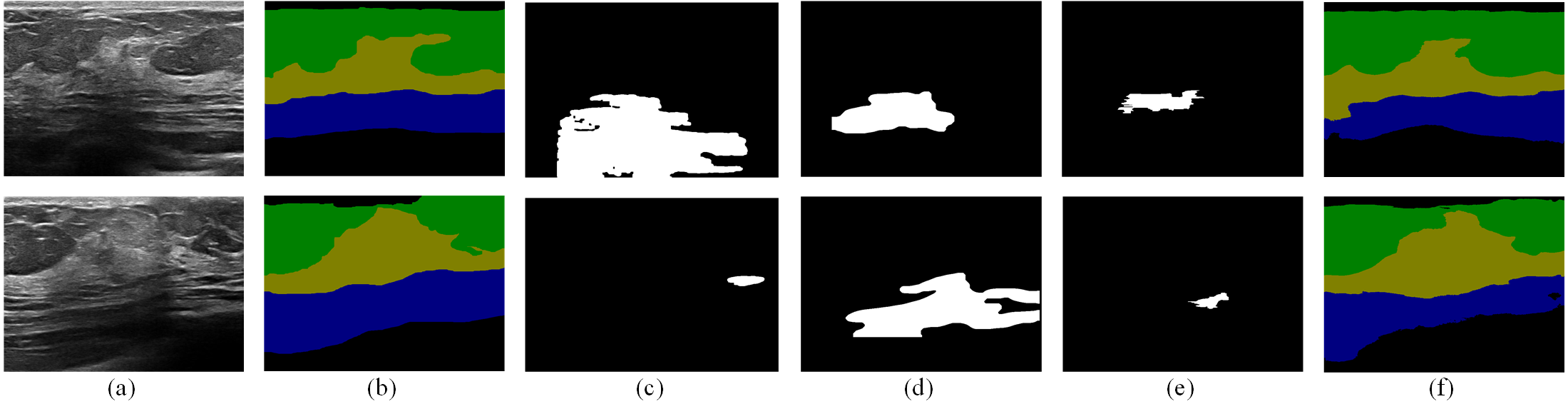}}
    \caption{The segmentation results for BUS images without tumors: (a) original images; (b) ground truths; (c) results using [34]; (d) results using [33]; (e) results using [35]; (f) results of the proposed method.}
\end{figure*}
\begin{table}[htbp]
\begin{center}
\caption{Evaluation Results on Tumor Segmentation}
\begin{tabular}{c|ccc}
\hline
 & TPR & FPR & IoU \\\hline
 & \multicolumn{3}{c}{Semi-Automatic Method} \\\hline
 Method \cite{ref31} & 83.01\% & 9.65\%	& 79.74\% \\
 Method \cite{ref9}	& 84.37\% & 17.51\% & 72.65\% \\\hline
 & \multicolumn{3}{c}{Fully-Automatic Method} \\\hline
 Method \cite{ref34} & 75.94\% & 43.84\% & 63.94\% \\
 Method \cite{ref33} & 83.55\% & 83.28\% & 65.22\% \\
 Method \cite{ref35} & 78.05\% & 15.43\% & 73.45\% \\
 Proposed & \textbf{90.33\%}	& \textbf{9.00\%} & \textbf{81.29\%} \\\hline
\end{tabular}
\label{tab3}
\end{center}
\end{table}

Table III shows that the proposed method achieves the best result among the methods in comparison. Furthermore, the proposed method can process the BUS images without tumors. The previous fully automatic methods could not solve such problem; since all of the previous methods are based on the prerequisite that there is one and only one tumor in the image \cite{ref16}. As shown in Fig. 18, the two samples do not contain tumors. Fig. 18 (c)-(e) are the results of the previous fully automatic methods \cite{ref33,ref34,ref35}, the white areas in the results are the tumors by the three methods, i.e., they do not work well.

From Fig. 18 (f), the proposed method can classify the layers in the BUS images well. It can provide medical context. After locating the layers in the breast, the anatomy of breast can be known which will be beneficial to tumor detection and segmentation.

Existing fully automatic segmentation methods \cite{ref33,ref34,ref35} cannot solve multi-tumor cases as well. In Fig. 19, three BUS images are not in our dataset, and each image contains 2 tumors. The first image is collected by a doctor of the First Affiliated Hospital of Harbin Medical University; the second one is found in a public dataset \cite{ref37}; and the third one is found in \cite{ref38}.

In Fig. 19, the existing methods (Fig. 19 (c-e)) can only detect one tumor for each image, i.e. they cannot obtain good results for containing more than one tumor; however, the proposed method can (Fig. 19 (f)).
\begin{figure}[tbp]
    \centerline{\includegraphics[width=8.5cm,height=4cm]{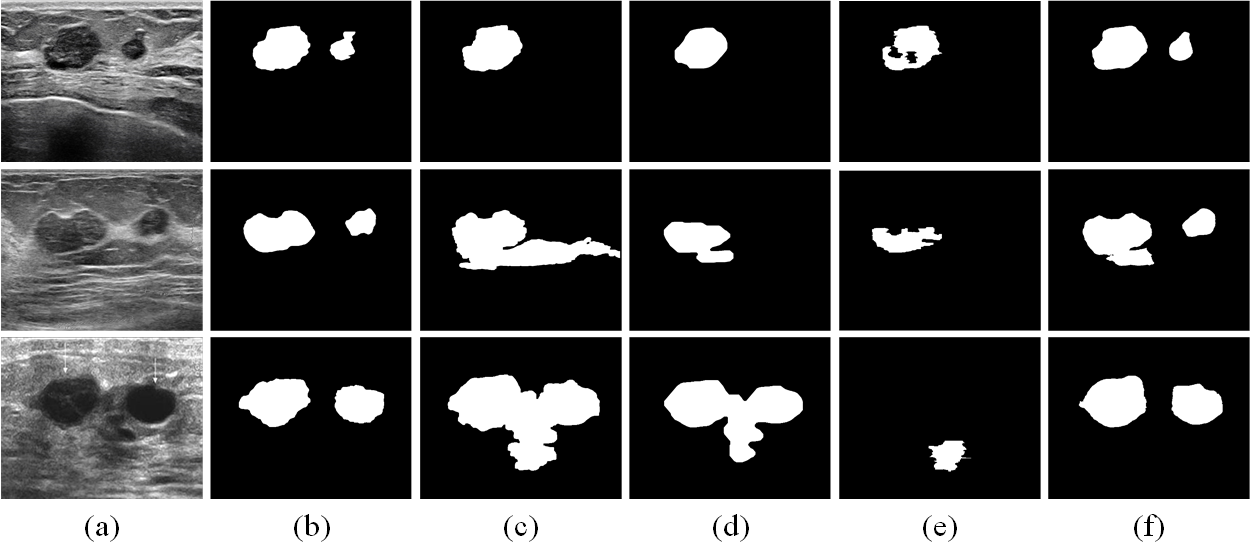}}
    \caption{The segmentation results of the BUS images containing two tumors: (a) the original images; (b) the ground truths; (c) results using [34]; (d) results using [33]; (e) results using [35]; (f) results using the proposed method}
\end{figure}
\section{Conclusion}
In this paper, a novel BUS image semantic segmentation method is proposed. It can achieve good semantic segmentation result. The approach consists of two steps. First, the fuzzy FCN can achieve good segmentation result. The second step uses breast anatomy constrained conditional random fields to fine-tune the segmentation result. The experimental results demonstrate that the proposed fuzzy FCN can handle the uncertainty well. The robustness and accuracy of the fuzzy FCN are better than that of the non-fuzzy FCN. 

The proposed method solves the following issues to achieve much better results: 1) it uses fuzzy logic to handle the uncertainty in the original image and feature maps of the convolutional layers; 2) fuzzy approach can provide more information; 3) it also provides anatomy information to fully connected CRFs which can increase the segmentation accuracy. There are still two potential improvements: 1) using fuzzy logic to handle the uncertainty in other convolutional layers and loss function; and 2) the anatomy context model of human breast is very complex; therefore, more anatomy information should be included.


%





\ifCLASSOPTIONcaptionsoff
  \newpage
\fi



\bibliographystyle{IEEEtran}
\bibliography{IEEEabrv,./bib/mybibfile}

\begin{thebibliography}{10}
\providecommand{\url}[1]{#1}
\csname url@samestyle\endcsname
\providecommand{\newblock}{\relax}
\providecommand{\bibinfo}[2]{#2}
\providecommand{\BIBentrySTDinterwordspacing}{\spaceskip=0pt\relax}
\providecommand{\BIBentryALTinterwordstretchfactor}{4}
\providecommand{\BIBentryALTinterwordspacing}{\spaceskip=\fontdimen2\font plus
\BIBentryALTinterwordstretchfactor\fontdimen3\font minus
  \fontdimen4\font\relax}
\providecommand{\BIBforeignlanguage}[2]{{%
\expandafter\ifx\csname l@#1\endcsname\relax
\typeout{** WARNING: IEEEtran.bst: No hyphenation pattern has been}%
\typeout{** loaded for the language `#1'. Using the pattern for}%
\typeout{** the default language instead.}%
\else
\language=\csname l@#1\endcsname
\fi
#2}}
\providecommand{\BIBdecl}{\relax}
\BIBdecl

\bibitem{ref1}
H.~D. Cheng, J.~Shan, W.~Ju, Y.~Guo, and L.~Zhang, ``Automated breast cancer
  detection and classification using ultrasound images: A survey,''
  \emph{Pattern Recognition}, vol.~43, no.~1, pp. 299--317, 2010.

\bibitem{ref2}
\BIBentryALTinterwordspacing
{\relax American Cancer Society}. (2019) How common is breast cancer. [Online].
  Available:
  \url{https://www.cancer.org/cancer/breast-cancer/about/how-common-is-breast-cancer.html}
\BIBentrySTDinterwordspacing

\bibitem{ref3}
C.~E. DeSantis, S.~A. Fedewa, A.~Goding~Sauer, J.~L. Kramer, R.~A. Smith, and
  A.~Jemal, ``Breast cancer statistics, 2015: Convergence of incidence rates
  between black and white women,'' \emph{CA: a cancer journal for clinicians},
  vol.~66, no.~1, pp. 31--42, 2016.

\bibitem{ref4}
Q.~Huang, Y.~Luo, and Q.~Zhang, ``Breast ultrasound image segmentation: a
  survey,'' \emph{International journal of computer assisted radiology and
  surgery}, vol.~12, no.~3, pp. 493--507, 2017.

\bibitem{ref30}
M.~{Xian}, Y.~{Zhang}, H.~D. {Cheng}, F.~{Xu}, K.~{Huang}, B.~{Zhang},
  J.~{Ding}, C.~{Ning}, and Y.~{Wang}, ``{A Benchmark for Breast Ultrasound
  Image Segmentation (BUSIS)},'' \emph{arXiv e-prints}, p. arXiv:1801.03182,
  Jan 2018.

\bibitem{ref36}
M.~Xian, Y.~Zhang, H.~D. Cheng, F.~Xu, B.~Zhang, and J.~Ding, ``Automatic
  breast ultrasound image segmentation: A survey,'' \emph{Pattern Recognition},
  vol.~79, pp. 340--355, 2018.

\bibitem{ref5}
Q.~Huang, F.~Yang, L.~Liu, and X.~Li, ``Automatic segmentation of breast
  lesions for interaction in ultrasonic computer-aided diagnosis,''
  \emph{Information Sciences}, vol. 314, pp. 293--310, 2015.

\bibitem{ref6}
J.~Z. Cheng, D.~Ni, Y.~H. Chou, J.~Qin, C.~M. Tiu, Y.~C. Chang, C.~S. Huang,
  D.~Shen, and C.~M. Chen, ``Computer-aided diagnosis with deep learning
  architecture: applications to breast lesions in us images and pulmonary
  nodules in ct scans,'' \emph{Scientific reports}, vol.~6, p. 24454, 2016.

\bibitem{ref7}
W.~K. Moon, C.~M. Lo, R.~T. Chen, Y.~W. Shen, J.~M. Chang, C.~S. Huang, J.~H.
  Chen, W.~W. Hsu, and R.~F. Chang, ``Tumor detection in automated breast
  ultrasound images using quantitative tissue clustering,'' \emph{Medical
  physics}, vol.~41, no.~4, p. 042901, 2014.

\bibitem{ref8}
J.~Shan, H.~D. Cheng, and Y.~Wang, ``A novel segmentation method for breast
  ultrasound images based on neutrosophic l-means clustering,'' \emph{Medical
  physics}, vol.~39, no.~9, pp. 5669--5682, 2012.

\bibitem{ref32}
M.~{Xian}, H.~D. {Cheng}, and Y.~{Zhang}, ``A fully automatic breast ultrasound
  image segmentation approach based on neutro-connectedness,'' in \emph{2014
  22nd International Conference on Pattern Recognition}, Aug 2014, pp.
  2495--2500.

\bibitem{ref9}
B.~Liu, H.~D. Cheng, J.~Huang, J.~Tian, X.~Tang, and J.~Liu, ``Probability
  density difference-based active contour for ultrasound image segmentation,''
  \emph{Pattern Recognition}, vol.~43, no.~6, pp. 2028--2042, 2010.

\bibitem{ref42}
M.~{Xian}, J.~{Huang}, Y.~{Zhang}, and X.~{Tang}, ``Multiple-domain knowledge
  based mrf model for tumor segmentation in breast ultrasound images,'' in
  \emph{2012 19th IEEE International Conference on Image Processing}, Sep.
  2012, pp. 2021--2024.

\bibitem{ref43}
M.~H. {Khan}, ``Automated breast cancer diagnosis using artificial neural
  network (ann),'' in \emph{2017 3rd Iranian Conference on Intelligent Systems
  and Signal Processing (ICSPIS)}, Dec 2017, pp. 54--58.

\bibitem{ref13}
M.~H. {Yap}, G.~{Pons}, J.~{Martí}, S.~{Ganau}, M.~{Sentís}, R.~{Zwiggelaar},
  A.~K. {Davison}, and R.~{Martí}, ``Automated breast ultrasound lesions
  detection using convolutional neural networks,'' \emph{IEEE Journal of
  Biomedical and Health Informatics}, vol.~22, no.~4, pp. 1218--1226, July
  2018.

\bibitem{ref16}
K.~{Huang}, H.~D. {Cheng}, Y.~{Zhang}, B.~{Zhang}, P.~{Xing}, and C.~{Ning},
  ``Medical knowledge constrained semantic breast ultrasound image
  segmentation,'' in \emph{2018 24th International Conference on Pattern
  Recognition (ICPR)}, Aug 2018, pp. 1193--1198.

\bibitem{ref17}
M.~Xian, Y.~Zhang, and H.~D. Cheng, ``Fully automatic segmentation of breast
  ultrasound images based on breast characteristics in space and frequency
  domains,'' \emph{Pattern Recognition}, vol.~48, no.~2, pp. 485--497, 2015.

\bibitem{ref10}
H.~Chen, Q.~Dou, X.~Wang, J.~Qin, and P.~A. Heng, ``Mitosis detection in breast
  cancer histology images via deep cascaded networks,'' in \emph{Thirtieth AAAI
  Conference on Artificial Intelligence}, 2016, pp. 1167–--1173.

\bibitem{ref11}
J.~{Liu}, B.~{Xu}, C.~{Zheng}, Y.~{Gong}, J.~{Garibaldi}, D.~{Soria},
  A.~{Green}, I.~O. {Ellis}, W.~{Zou}, and G.~{Qiu}, ``An end-to-end deep
  learning histochemical scoring system for breast cancer tma,'' \emph{IEEE
  Transactions on Medical Imaging}, vol.~38, no.~2, pp. 617--628, Feb 2019.

\bibitem{ref12}
W.~{Zhu}, X.~{Xiang}, T.~D. {Tran}, G.~D. {Hager}, and X.~{Xie}, ``Adversarial
  deep structured nets for mass segmentation from mammograms,'' in \emph{2018
  IEEE 15th International Symposium on Biomedical Imaging (ISBI 2018)}, April
  2018, pp. 847--850.

\bibitem{ref41}
Y.~{Lecun}, L.~{Bottou}, Y.~{Bengio}, and P.~{Haffner}, ``Gradient-based
  learning applied to document recognition,'' \emph{Proceedings of the IEEE},
  vol.~86, no.~11, pp. 2278--2324, Nov 1998.

\bibitem{ref14}
O.~Ronneberger, P.~Fischer, and T.~Brox, ``U-net: Convolutional networks for
  biomedical image segmentation,'' in \emph{Medical Image Computing and
  Computer-Assisted Intervention -- MICCAI 2015}, N.~Navab, J.~Hornegger, W.~M.
  Wells, and A.~F. Frangi, Eds.\hskip 1em plus 0.5em minus 0.4em\relax Cham:
  Springer International Publishing, 2015, pp. 234--241.

\bibitem{ref15}
A.~Krizhevsky, I.~Sutskever, and G.~E. Hinton, ``Imagenet classification with
  deep convolutional neural networks,'' in \emph{Advances in Neural Information
  Processing Systems 25}, 2012, pp. 1097--1105.

\bibitem{ref18}
Y.~{Deng}, Z.~{Ren}, Y.~{Kong}, F.~{Bao}, and Q.~{Dai}, ``A hierarchical fused
  fuzzy deep neural network for data classification,'' \emph{IEEE Transactions
  on Fuzzy Systems}, vol.~25, no.~4, pp. 1006--1012, Aug 2017.

\bibitem{ref20}
G.~N. Sharma, R.~Dave, J.~Sanadya, P.~Sharma, and K.~Sharma, ``Various types
  and management of breast cancer: an overview,'' \emph{Journal of advanced
  pharmaceutical technology \& research}, vol.~1, no.~2, pp. 109--126, 2010.

\bibitem{ref21}
J.~{Long}, E.~{Shelhamer}, and T.~{Darrell}, ``Fully convolutional networks for
  semantic segmentation,'' in \emph{2015 IEEE Conference on Computer Vision and
  Pattern Recognition (CVPR)}, June 2015, pp. 3431--3440.

\bibitem{ref40}
{Yeong-Taeg Kim}, ``Contrast enhancement using brightness preserving
  bi-histogram equalization,'' \emph{IEEE Transactions on Consumer
  Electronics}, vol.~43, no.~1, pp. 1--8, Feb 1997.

\bibitem{ref23}
H.~D. Cheng and H.~Xu, ``A novel fuzzy logic approach to contrast
  enhancement,'' \emph{Pattern recognition}, vol.~33, no.~5, pp. 809--819,
  2000.

\bibitem{ref24}
K.~S. Chuang, H.~L. Tzeng, S.~Chen, J.~Wu, and T.~J. Chen, ``Fuzzy c-means
  clustering with spatial information for image segmentation,''
  \emph{computerized medical imaging and graphics}, vol.~30, no.~1, pp. 9--15,
  2006.

\bibitem{ref25}
P.~{Melin}, C.~I. {Gonzalez}, J.~R. {Castro}, O.~{Mendoza}, and O.~{Castillo},
  ``Edge-detection method for image processing based on generalized type-2
  fuzzy logic,'' \emph{IEEE Transactions on Fuzzy Systems}, vol.~22, no.~6, pp.
  1515--1525, Dec 2014.

\bibitem{ref28}
P.~Kr{\"a}henb{\"u}hl and V.~Koltun, ``Efficient inference in fully connected
  crfs with gaussian edge potentials,'' in \emph{Advances in neural information
  processing systems}, 2011, pp. 109--117.

\bibitem{ref26}
L.~C. {Chen}, G.~{Papandreou}, I.~{Kokkinos}, K.~{Murphy}, and A.~L. {Yuille},
  ``Deeplab: Semantic image segmentation with deep convolutional nets, atrous
  convolution, and fully connected crfs,'' \emph{IEEE Transactions on Pattern
  Analysis and Machine Intelligence}, vol.~40, no.~4, pp. 834--848, April 2018.

\bibitem{ref27}
L.~C. Chen, G.~Papandreou, I.~Kokkinos, K.~Murphy, and A.~L. Yuille, ``Semantic
  image segmentation with deep convolutional nets and fully connected crfs,''
  in \emph{Proceedings of International Conference on Learning
  Representations}, 2015.

\bibitem{ref22}
S.~Zheng, S.~Jayasumana, B.~Romera-Paredes, V.~Vineet, Z.~Su, D.~Du, C.~Huang,
  and P.~H. Torr, ``Conditional random fields as recurrent neural networks,''
  in \emph{Proceedings of the IEEE international conference on computer
  vision}, 2015, pp. 1529--1537.

\bibitem{ref19}
Z.~Liu, X.~Li, P.~Luo, C.~C. Loy, and X.~Tang, ``Semantic image segmentation
  via deep parsing network,'' in \emph{Proceedings of the IEEE international
  conference on computer vision}, 2015, pp. 1377--1385.

\bibitem{ref39}
D.~Ramsay, J.~Kent, R.~Hartmann, and P.~Hartmann, ``Anatomy of the lactating
  human breast redefined with ultrasound imaging,'' \emph{Journal of anatomy},
  vol. 206, no.~6, pp. 525--534, 2005.

\bibitem{ref29}
A.~Adams, J.~Baek, and M.~A. Davis, ``Fast high-dimensional filtering using the
  permutohedral lattice,'' in \emph{Computer Graphics Forum}, vol.~29,
  no.~2.\hskip 1em plus 0.5em minus 0.4em\relax Wiley Online Library, 2010, pp.
  753--762.

\bibitem{ref31}
Y.~Liu, H.~D. Cheng, J.~Huang, Y.~Zhang, and X.~Tang, ``An effective approach
  of lesion segmentation within the breast ultrasound image based on the
  cellular automata principle,'' \emph{Journal of digital imaging}, vol.~25,
  no.~5, pp. 580--590, 2012.

\bibitem{ref33}
M.~Xian, Y.~Zhang, and H.~D. Cheng, ``Fully automatic segmentation of breast
  ultrasound images based on breast characteristics in space and frequency
  domains,'' \emph{Pattern Recognition}, vol.~48, no.~2, pp. 485--497, 2015.

\bibitem{ref34}
J.~Shan, H.~D. Cheng, and Y.~Wang, ``Completely automated segmentation approach
  for breast ultrasound images using multiple-domain features,''
  \emph{Ultrasound in medicine \& biology}, vol.~38, no.~2, pp. 262--275, 2012.

\bibitem{ref35}
H.~{Shao}, Y.~{Zhang}, M.~{Xian}, H.~D. {Cheng}, F.~{Xu}, and J.~{Ding}, ``A
  saliency model for automated tumor detection in breast ultrasound images,''
  in \emph{2015 IEEE International Conference on Image Processing (ICIP)}, Sep.
  2015, pp. 1424--1428.

\bibitem{ref37}
A.~Rodtook and S.~S. Makhanov, ``Multi-feature gradient vector flow snakes for
  adaptive segmentation of the ultrasound images of breast cancer,''
  \emph{Journal of Visual Communication and Image Representation}, vol.~24,
  no.~8, pp. 1414--1430, 2013.

\bibitem{ref38}
\BIBentryALTinterwordspacing
{M. Alice}. (2018) Breast ultrasound cancer vs benign. [Online]. Available:
  \url{https://canceroz.blogspot.com/2017/02/breast-ultrasound-cancer-vs-benign.html}
\BIBentrySTDinterwordspacing

\end{thebibliography}
%


%
\begin{IEEEbiography}[{\includegraphics[width=1in,height=1.25in,clip,keepaspectratio]{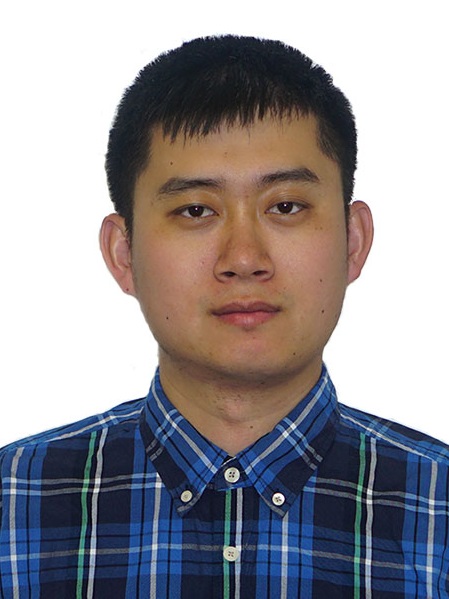}}]{Kuan Huang}
received his B.Eng. degree from the School of Electrical Engineering \& Automation of Harbin Institute of Technology, Harbin, China, in 2016. He is now a Ph.D. student in the Department of Computer Science, Utah State University. His current research interests include computer vision, machine learning, pattern recognition, medical image analysis, and fuzzy logic. 
\end{IEEEbiography}


\begin{IEEEbiography}[{\includegraphics[width=1in,height=1.25in,clip,keepaspectratio]{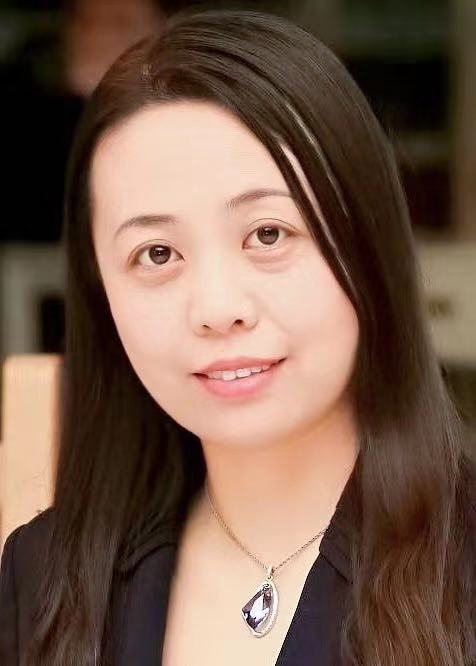}}]{Yingtao Zhang}
received her M.S. degree from Computer Science School of Harbin Institute of Technology, Harbin, China, in 2004, and the Ph.D. degree in Pattern Recognition and Intelligence System from Harbin Institute of Technology, Harbin, China, in 2010. Now, she is an associate professor, School of Computer Science and Technology, Harbin Institute of Technology, Harbin, China. Her research interests include pattern recognition, computer vision, and medical image processing.
\end{IEEEbiography}
\begin{IEEEbiography}[{\includegraphics[width=1in,height=1.25in,clip,keepaspectratio]{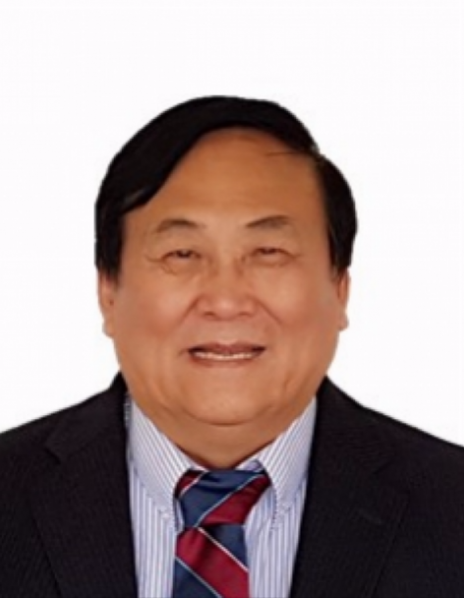}}]{H. D. Cheng}
received the Ph.D. degree in Electrical Engineering from Purdue University (Supervisor: Prof. K.S. Fu), West Lafayette, IN, 1985. Now, he is a Full Professor, Department of Computer Science, and an Adjunct Full Professor, Department of Electrical Engineering, Utah State University, Logan, Utah. Dr.  Cheng is an Adjunct Professor and Doctorial Supervisor of Harbin Institute of Technology. He is also a Guest Professor of the Institute of Remote Sensing Application, Chinese Academy of Sciences, a Guest Professor of Wuhan University, a Guest Professor of Shantou University, and a Visiting Professor of Northern Jiaotong University.

Dr. Cheng has published more than 350 technical papers and is the Co-editor of the book, Pattern Recognition: Algorithms, Architectures and Applications (World Scientiﬁc Publishing Co., 1991).
His research interests include image processing, pattern recognition, computer vision, artiﬁcial intelligence, medical information processing, fuzzy logic, genetic algorithms, neural networks, parallel processing, parallel algorithms, and VLSI architectures.

Dr. Cheng was the General Chair of the 11th Joint Conference on Information Sciences (JCIS 2008), the General Chair of the 10th  Joint Conference on Information Sciences (JCIS 2007), the General Chair of the Ninth Joint Conference on Information Sciences (JCIS 2006), the General Chair of the Eighth Joint Conference on Information Sciences (JCIS 2005), etc. He served as program committee member and Session Chair for many conferences, and as reviewer for many scientiﬁc journals and conferences. Dr. Cheng has been listed in Who’s Who in the World, Who’s Who in America, Who’s Who in Communications and Media, etc.

Dr. Cheng is also an Associate Editor of Pattern Recognition, an Associate Editor of Information Sciences and Associate Editor of New Mathematics and Natural Computation.
\end{IEEEbiography}
\begin{IEEEbiography}[{\includegraphics[width=1in,height=1.25in,clip,keepaspectratio]{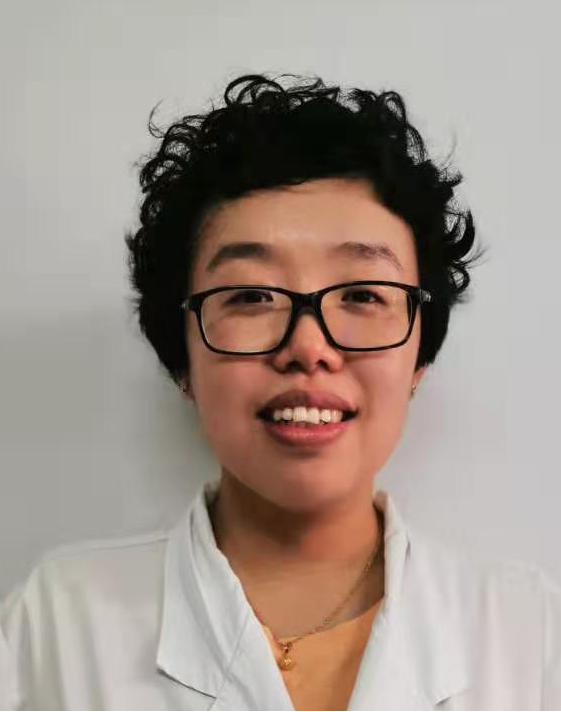}}]{Ping Xing}
received her Ph.D. degree from Harbin Medical University, Harbin, China, in 2011. She is major in medical imaging. Now, she is working in the Department of Ultrasound, the First Hospital of Harbin Medical University, Harbin, China. Her research interests include ultrasound diagnosis of the thyroid diseases, and breast diseases.
\end{IEEEbiography}
\begin{IEEEbiography}[{\includegraphics[width=1in,height=1.25in,clip,keepaspectratio]{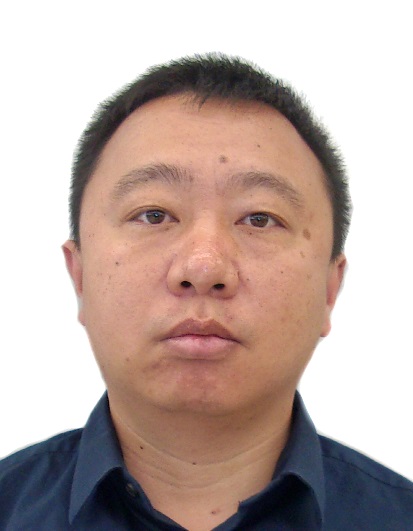}}]{Boyu Zhang}
received his M.S. and Ph.D. degrees in computer science from the School of Computer Science in Harbin Institute of Technology, Harbin, China, in 2009, and 2015, respectively. Now, he is a post-doctoral research fellow in the Department of Computer Science, Utah State University, Logan, Utah. His research interests include pattern recognition, computer vision, and medical image.
\end{IEEEbiography}




\end{document}